%% file: techreport.tex
\documentclass{agibot-report}
\input{math_commands.tex}

\usepackage{placeins}
\usepackage{amssymb}
\usepackage{booktabs}
\usepackage{multirow}
\usepackage{caption}
\usepackage{microtype}
\usepackage{xspace}
\usepackage{algorithm}
\usepackage{algpseudocode}

\newcommand{\modelname}{Rollback\xspace}
\definecolor{rbgain}{HTML}{2E8B6E}

\renewenvironment{abstract}{%
  \vskip 0.06in
  \tcolorbox[enhanced, colback=niceblue!7!white, colframe=niceblue!7!white,
    boxrule=0pt, arc=3pt, outer arc=3pt, left=14pt, right=14pt, top=7pt, bottom=9pt,
    enlarge left by=18pt, enlarge right by=18pt, width=\linewidth-36pt]
  \centerline{\large\bfseries Abstract}\vspace{0.7ex}}
 {\endtcolorbox\vskip 1.2ex}

\DeclareCaptionFormat{loose}{\looseness=-1 #1#2#3\par}
\title{\raisebox{-6.5pt}{\includegraphics[height=25pt]{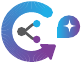}}\hspace{4pt}%
  Rollback: Experience Backtracking in World Models for Policy Improvement}
\hypersetup{pdftitle={Rollback: Experience Backtracking in World Models for Policy Improvement}}

\newcommand{\af}[1]{\textsuperscript{\normalfont #1}}
\newcommand{\asep}{\hspace{1.4em}}
\author{%
  Jiayi~Luo\af{1,2,$\ast$}\asep
  Yang~Zhang\af{1,$\ast$}\asep
  Liliang~Chen\af{1,$\ast$,$\dagger$}\asep
  Nan~Wang\af{1}\asep
  Chen~Gao\af{2,3}\asep
  Boxiang~Qiu\af{1,2}\\
  Zhuoxuan~Li\af{1}\asep
  Baorui~Peng\af{1}\asep
  Chuankang~Li\af{1}\asep
  Yuxiang~Yan\af{1}\asep
  Si~Liu\af{2}\asep
  Qingyun~Sun\af{2}\\
  Shuicheng~Yan\af{3}\asep
  Maoqing~Yao\af{1}\asep
  Jianxin~Li\af{2,$\diamond$}\asep
  Guanghui~Ren\af{1,$\diamond$}}
\affiliation{\bfseries \af{1}\,AgiBot \qquad \af{2}\,BUAA \qquad \af{3}\,NUS}
\authornote{$^\ast$\,Equal Contribution. \quad $^\dagger$\,Project Leader. \quad $^\diamond$\,Corresponding Author.}
\projectpage{https://rollback-wm.github.io/}

\begin{document}

\maketitle

\input{techreport/section/0_abstract}
\input{techreport/section/1_introduction}
\input{techreport/section/2_related}
\input{techreport/section/3_preliminaries}
\input{techreport/section/4_method}
\input{techreport/section/5_system}
\input{techreport/section/6_experiment}
\input{techreport/section/7_conclusion}

\bibliography{iclr2027_conference}

\input{techreport/section/8_appendix}

\end{document}

%% file: math_commands.tex
\usepackage{amsmath,amsfonts,bm}

\def\eqref#1{equation~\ref{#1}}

\def\1{\bm{1}}

\DeclareMathAlphabet{\mathsfit}{\encodingdefault}{\sfdefault}{m}{sl}
\SetMathAlphabet{\mathsfit}{bold}{\encodingdefault}{\sfdefault}{bx}{n}

%% file: techreport/section/0_abstract.tex
\begin{abstract}

Action-conditioned world models offer a scalable alternative to costly real-world interaction for improving robot policies.
However, existing world-model-based methods largely treat world models as forward-only surrogates of the physical environment, \textit{leaving untapped a capability no physical environment affords: rewinding to past decision points to explore counterfactual futures}.
In this work, we introduce \textbf{\modelname}, a policy improvement paradigm that turns world models from passive simulators into counterfactual engines: it rolls back to critical decision points, branches into alternative actions, and contrasts their outcomes to attribute success to individual decisions.
This concentrates imagination where it matters most and distills counterfactual outcomes into fine-grained, action-level supervision without a separately learned value model.
Furthermore, we build a scalable training system that unifies distributed world-model search with asynchronous real-robot execution, forming a self-reinforcing cycle: real-world experience continually sharpens the world model, and a sharper world model in turn drives further policy improvement.
We term the world-model side of this cycle \emph{test-time physics grounding}, a capability that hand-built physics simulators lack: while the world model serves as the training environment, real transitions from the policy it trains correct its dynamics through a lightweight LoRA pipeline.
Experiments on three challenging real-world tasks show that \modelname improves success rates over the strongest baseline by an absolute 10.67\%, 11.11\%, and 11.85\% on block extraction, ball tossing, and tabletop curling, respectively, while achieving more stable improvement throughout training.
Although initialized from teleoperated demonstrations far below human level, \modelname surpasses the average success rate of human players on both ball tossing and tabletop curling within 8 hours of training.
\end{abstract}

%% file: techreport/section/1_introduction.tex
\section{Introduction}
\label{sec:intro}

Action-conditioned world models have provided simulated environments for robot
policy training, offering a scalable complement to real-world
interaction~\citep{planet,dreamer,dreamerv2}.
By predicting the consequences of policy-generated actions, they allow policies
to collect experience and learn from imagined outcomes without physical
deployment~\citep{unisim,irasim,enerverseac}.
Recent video world models extend this paradigm to diverse robotic tasks,
supporting policy evaluation~\citep{worldeval,veorobotics} and improvement through
synthetic experience~\citep{dreamgen,rise,interactiveworldsim}.
These advances further extend to general-purpose robot policies, which can be
refined using imagined interactions and task feedback~\citep{worldenv,wmpo}.
Moving exploration and optimization into learned environments makes policy
learning less bound to physical trials.

Despite this progress, \textit{existing world-model-based policy learning
largely retains the trajectory-centric interaction protocol of real-world training},
using the world model primarily to generate forward rollouts whose
intermediate contexts are discarded afterward.
This overlooks a capability unavailable in physical RL: the ability to
``rewind and reconsider.''
While a physical environment cannot generally return to an identical
intermediate state after execution, a world model can restore a cached temporal
context, rewind to an earlier decision point, and explore alternative action
continuations from the same history.
This enables the policy to reconsider what it could have done differently at a
particular decision rather than restarting exploration from scratch.

\begin{figure}[t]
  \centering
  \includegraphics[width=\linewidth]{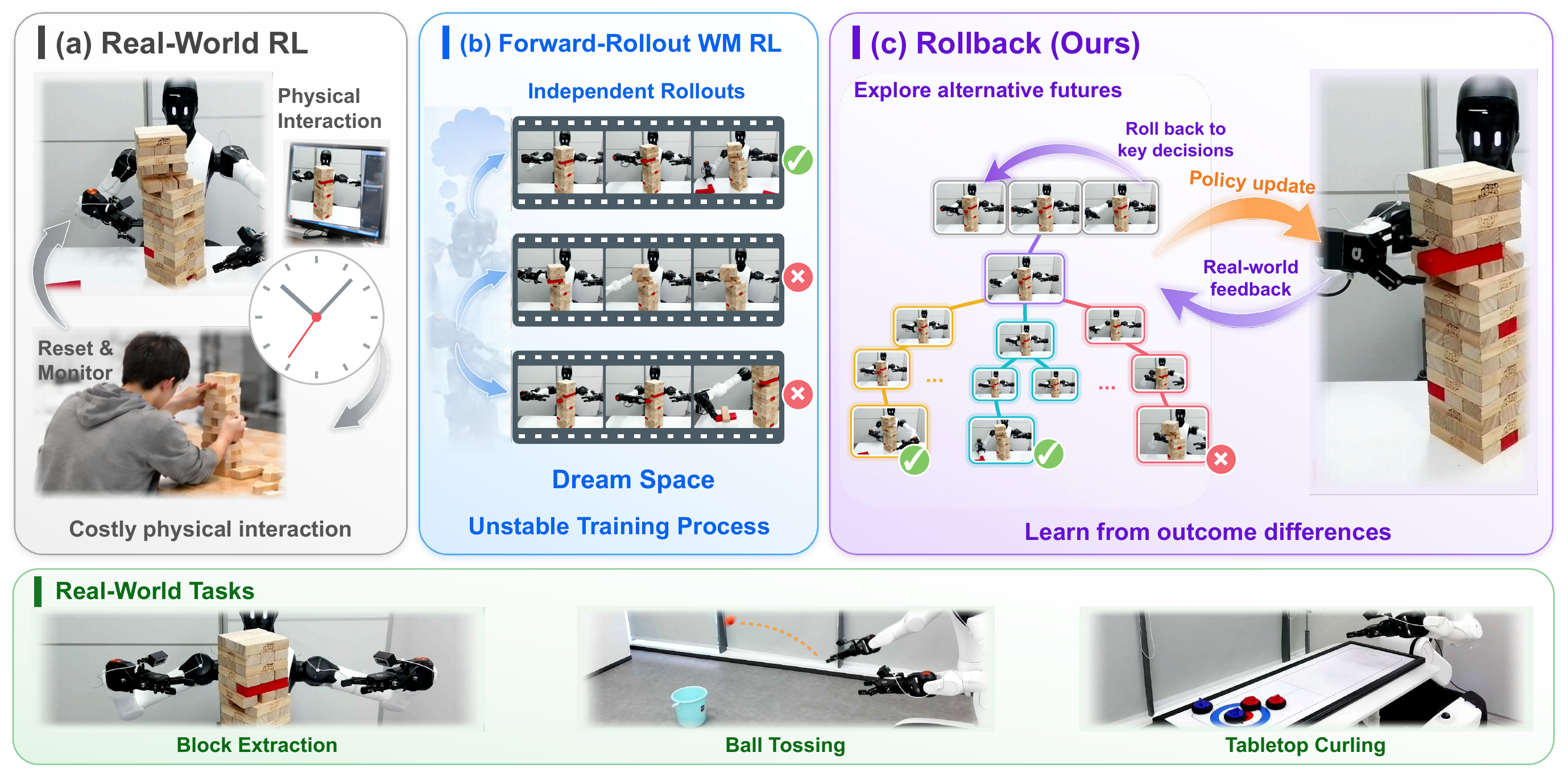}
    \captionsetup{format=loose}
\caption{
\textbf{From real-world interaction to learning in world models.}
(a) Real-world RL incurs costly physical interaction.
(b) Forward-rollout world-model RL spends much of its interaction budget on less informative exploration.
(c) Our proposed \modelname enables high-value exploration and dense supervision
through experience-backtracking tree search, while real-world feedback refines the world model.
}
  \label{fig:teaser}
\end{figure}

Fully exploiting this capability, however, requires addressing three key
challenges as shown in Figure~\ref{fig:teaser}.
\textit{(i) Limited informative exploration.}
Not all decisions are equally worth revisiting: along a trajectory, a few critical decisions determine the outcome, while alternatives to most others lead to similar results.
Independent rollouts cannot identify these consequential decisions or adaptively allocate exploration among them~\citep{goexplore,mcts}.
\textit{(ii) Sparse decision-level supervision.}
Trajectory-level feedback does not reveal which intermediate actions determine the outcome~\citep{her,riptvla}.
Treating continuations independently preserves this credit-assignment problem, whereas learning an additional value model introduces extra modeling assumptions and potential estimation bias~\citep{vlarl,gae,td3}.
\textit{(iii) Biased world-model predictions.}
Branching from policy-selected contexts may shift interactions beyond the world model's training distribution, causing prediction errors to accumulate or be exploited during optimization~\citep{mbpo,mopo,combo,hawmpo}.
Without real-world grounding, imagined gains may not transfer to real robots.

\begin{figure}[t]
\begin{agibox}[title=Key insight, unbreakable]
\textbf{\textit{Our key insight is therefore to redesign world-model-based
policy learning around a training-time experience backtracking primitive}},
repeatedly rewinding to and reconsidering decisions for informative data
generation.
The world model becomes not merely a surrogate simulator, but a rewindable
training environment.
\end{agibox}
\end{figure}

\looseness=-1 To operationalize training-time experience backtracking and address these challenges, we present \textbf{\modelname}, a policy improvement paradigm that turns world models from passive simulators into counterfactual engines.
\modelname comprises three key components:
\textit{(i) Experience-backtracking tree search.}
Starting from successful and failed real-world trajectories, \modelname rolls back to intermediate decision points and branches into alternative action chunks.
Monte Carlo tree search adaptively allocates simulation budget toward decisions whose alternatives lead to divergent outcomes, avoiding exhaustive branching~\citep{mcts,alphago,sampledmuzero}.
\textit{(ii) Tree-structured value estimation.}
Since sibling branches share a decision context, contrasting their downstream outcomes isolates the effect of individual actions.
Subtree-level outcome aggregation yields fine-grained, action-level advantage supervision without an additional learned value model.
\textit{(iii) World model--policy co-evolution.}
A scalable training system unifies distributed world-model search with asynchronous real-robot execution.
Real-world experience continually refines the world model, which in turn provides more faithful imagination for search and policy improvement.
We call this \emph{test-time physics grounding}: after pretraining, while the world model serves as the training environment for a policy, real transitions collected by that policy correct its predicted dynamics in the states the policy actually visits.
Most existing action-conditioned world models are trained offline and then used as fixed simulators~\citep{unisim,irasim,enerverseac,ctrlworld}, which suffices for evaluation but not for policy learning, because an improving policy moves into exactly the states the simulator has seen least.

We evaluate \modelname on three challenging real-world robotics tasks: block extraction, ball tossing, and tabletop curling.
Across these tasks, \modelname improves success rates over the strongest baseline by an absolute 10.67\%, 11.11\%, and 11.85\%, respectively.
Under matched interaction budgets, \modelname achieves faster and more consistent improvements in task success rates.
Within 8 hours of training with one physical robot, \modelname surpasses the average success rate of human players on both ball tossing and tabletop curling, although it is initialized from teleoperated demonstrations far below human level.
Our project website is available at \href{https://rollback-wm.github.io/}{\texttt{rollback-wm.github.io}}.

In summary, our contributions in this work can be summarized as follows:
\begin{itemize}[leftmargin=*]
    \item We propose \modelname, a policy improvement paradigm that exploits experience backtracking to turn world models from passive simulators into counterfactual engines, supported by a training system unifying distributed world-model search with asynchronous real-robot execution.
    \item We design an experience-backtracking search that concentrates imagination on critical decision points, and a tree-structured value estimation that distills counterfactual outcomes into fine-grained, action-level supervision without a separately learned value model.
    \item Extensive experiments on three challenging real-world robotics tasks demonstrate that \modelname achieves higher success rates and more stable improvement.
\end{itemize}

%% file: techreport/section/2_related.tex
\section{Related Work}
\label{sec:related}

\paragraph{Robot Policy Learning from Interaction.}
Robot policies are commonly improved for downstream tasks and deployment conditions using task-specific data and interaction feedback~\citep{openvlaoft,irevla,recap}.
For generalist policies including RT-2, OpenVLA, and the $\pi$ series~\citep{rt2,openvla,pi0,pi05}, supervised fine-tuning learns task-specific behaviors; OpenVLA-OFT improves adaptation and inference efficiency through optimized action representations and objectives~\citep{openvlaoft}.
Reinforcement learning further exploits execution outcomes: iRe-VLA alternates reinforcement and supervised learning for stability~\citep{irevla}; VLA-RL learns a process reward model~\citep{vlarl}; RIPT-VLA uses dynamic sampling and leave-one-out advantages to learn from sparse success without a value network~\citep{riptvla}; and SimpleVLA-RL scales training through efficient sampling and parallelization~\citep{simplevlarl}.
For real-world adaptation, RECAP combines demonstrations, autonomous experience, and expert corrections through advantage-conditioned training~\citep{recap}.
\textit{However, reinforcement learning on physical robots remains resource-intensive due to costly interactions and labor-intensive environment resets}~\citep{rise,rlinfuser}.

\paragraph{World Models for Robotics.}
World models support robot learning and planning by predicting the consequences of candidate actions before physical execution~\citep{planet,dreamer,unisim}.
Recent advances span general-purpose world simulation and models tailored to robot interaction.
Cosmos-Predict2.5 and LingBot-World advance controllable video generation and interactive simulation~\citep{cosmospredict25,lingbotworld}.
UniSim, IRASim, EnerVerse-AC, and Ctrl-World model action-conditioned visual dynamics to support simulated interaction and policy development~\citep{unisim,irasim,enerverseac,ctrlworld}.
DreamDojo further transfers interaction knowledge from large-scale human videos to robot world modeling through latent actions~\citep{dreamdojo}.
WorldEval, WorldGym, and the Veo robotics simulator study generated rollouts as proxies for real-world policy evaluation~\citep{worldeval,worldgym,veorobotics}.
Beyond simulation, EnerVerse, Genie Envisioner, and LingBot-VA connect future visual prediction with action generation for robot control~\citep{enerverse,genie,lingbotva}.
DreamGen uses generated videos and inferred actions to construct synthetic training trajectories~\citep{dreamgen}, while V-JEPA 2-AC supports robotic planning through action-conditioned prediction in latent space~\citep{vjepa2}.

\paragraph{World-Model-Based Policy Improvement.}
\looseness=-1 Recent work increasingly uses action-conditioned world models as interactive environments for policy improvement~\citep{worldenv,wmpo,worldgymnast,rise,wovr}.
World-Env combines a video simulator with a VLM-guided reflector that supplies task feedback and termination signals~\citep{worldenv}, while WMPO optimizes policies through interactions with a learned world model~\citep{wmpo}.
World-Gymnast uses VLM-evaluated imagined rollouts for reinforcement learning and investigates iterative world-model updates using real-world experience~\citep{worldgymnast}.
To obtain informative supervision, RISE combines a controllable dynamics model with a learned progress value model that provides advantages for policy improvement~\citep{rise}.
WoVR instead emphasizes the reliability of imagined interaction through keyframe-initialized rollouts and world model--policy co-evolution~\citep{wovr}.
\textit{However, these methods struggle to improve steadily, as forward-only rollouts rarely revisit informative decisions and learned values may be biased.}

\paragraph{Physics Grounding of World Simulators.}
Action-conditioned world models are typically trained offline on large collections of robot or human videos and then evaluated or deployed as fixed simulators~\citep{unisim,irasim,enerverseac,ctrlworld,dreamdojo}.
Their fidelity is measured on held-out data from the training distribution, and how a simulator should remain physically grounded once a policy starts exploring it has received little attention.
Physics engines offer a contrast: their dynamics are specified by hand, and system identification can only tune the parameters of a fixed model.
A learned world simulator can instead absorb new real transitions directly, so its physics can be corrected while it is in use.
World-Gymnast and WoVR update the world model with real experience as part of their training loops~\citep{worldgymnast,wovr}.
We isolate this update as a capability of the simulator itself, test-time physics grounding, and show that it matters for steady policy improvement in world-model-based reinforcement learning.

\paragraph{Tree Search for Policy Improvement.}
Monte Carlo tree search allocates a limited simulation budget by balancing the exploitation of promising branches with the exploration of under-visited ones~\citep{mcts}.
AlphaGo and Expert Iteration distill search-improved decisions back into a policy network~\citep{alphago,expertiteration}, MuZero plans with a learned latent dynamics model~\citep{muzero}, and Sampled MuZero extends search to complex and continuous action spaces through action sampling~\citep{sampledmuzero}.
Go-Explore further shows that returning to previously visited states before exploring substantially improves exploration in hard-exploration problems~\citep{goexplore}.
\textit{However, these methods rely on exact simulators or compact latent models, leaving the ability of video world models to restore past contexts and branch counterfactual futures unexplored for real-world robot policy improvement.}

%% file: techreport/section/3_preliminaries.tex
\section{Preliminaries}
\label{sec:prelim}

\subsection{Problem Formulation}
\label{sec:problem}
Given a language instruction $g$ and an observation $o_t$ comprising multi-view images and proprioception, the policy $\pi_\theta$ generates an action chunk $\mathbf{a}_t\in\mathbb{R}^{H\times d_a}$ with horizon $H$ and action dimension $d_a$. The action-conditioned world model $\mathcal{W}_\phi$ predicts the next observation and updates its temporal context through $(\hat{o}_{t+1},m_{t+1})\sim\mathcal{W}_\phi(\cdot\mid m_t,\mathbf{a}_t,g)$, where $m_t$ denotes the retained temporal context. The predicted observation $\hat{o}_{t+1}$ contains both the multi-view images and the proprioceptive state, so the policy receives the same inputs in imagination as on the real robot. Our objective is to improve the policy's real-world task success by learning from imagined world-model interactions and real-world experience.
A real-world episode $\tau=(o_0,\mathbf{a}_0,o_1,\mathbf{a}_1,\ldots,o_T)$ ends with a binary outcome $y_\tau\in\{0,1\}$ indicating task success, so the objective amounts to maximizing $J(\pi_\theta)=\mathbb{E}_{\tau\sim\pi_\theta}[y_\tau]$ with as little physical interaction as possible.
Real episodes are expensive since each occupies the robot and needs a manual reset, whereas GPU compute scales with hardware.

\subsection{World Models as Rewindable Environments}
\label{sec:rewindable}
Unlike a physical environment, a world model can also be rewound.
Because $m_t$ is an explicit data structure, it can be cached and restored, so many continuations can branch from one decision point:
\begin{equation}
    m_t \;\xrightarrow{\;\textsc{Restore}\;}\;
    \bigl\{(\hat{o}^{(i)}_{t+1},m^{(i)}_{t+1})\sim\mathcal{W}_\phi(\cdot\mid m_t,\mathbf{a}^{(i)}_t,g)\bigr\}_{i=1}^{n}.
    \label{eq:restore}
\end{equation}
We refer to this operation as \emph{experience backtracking}.
It differs from resetting a simulator to its initial state in two respects.
First, the restored context can be anchored to real experience: conditioning $\mathcal{W}_\phi$ on a recorded observation starts imagination from a situation that the robot actually faced, including the decision points at which a real attempt succeeded or failed.
Second, the shared prefix is never regenerated, so the cost of a new continuation depends only on its own length rather than on the length of the full trajectory.

%% file: techreport/section/4_method.tex
\section{Proposed \modelname}
\label{sec:method}
\begin{figure}[t]
  \centering
  \includegraphics[width=\linewidth]{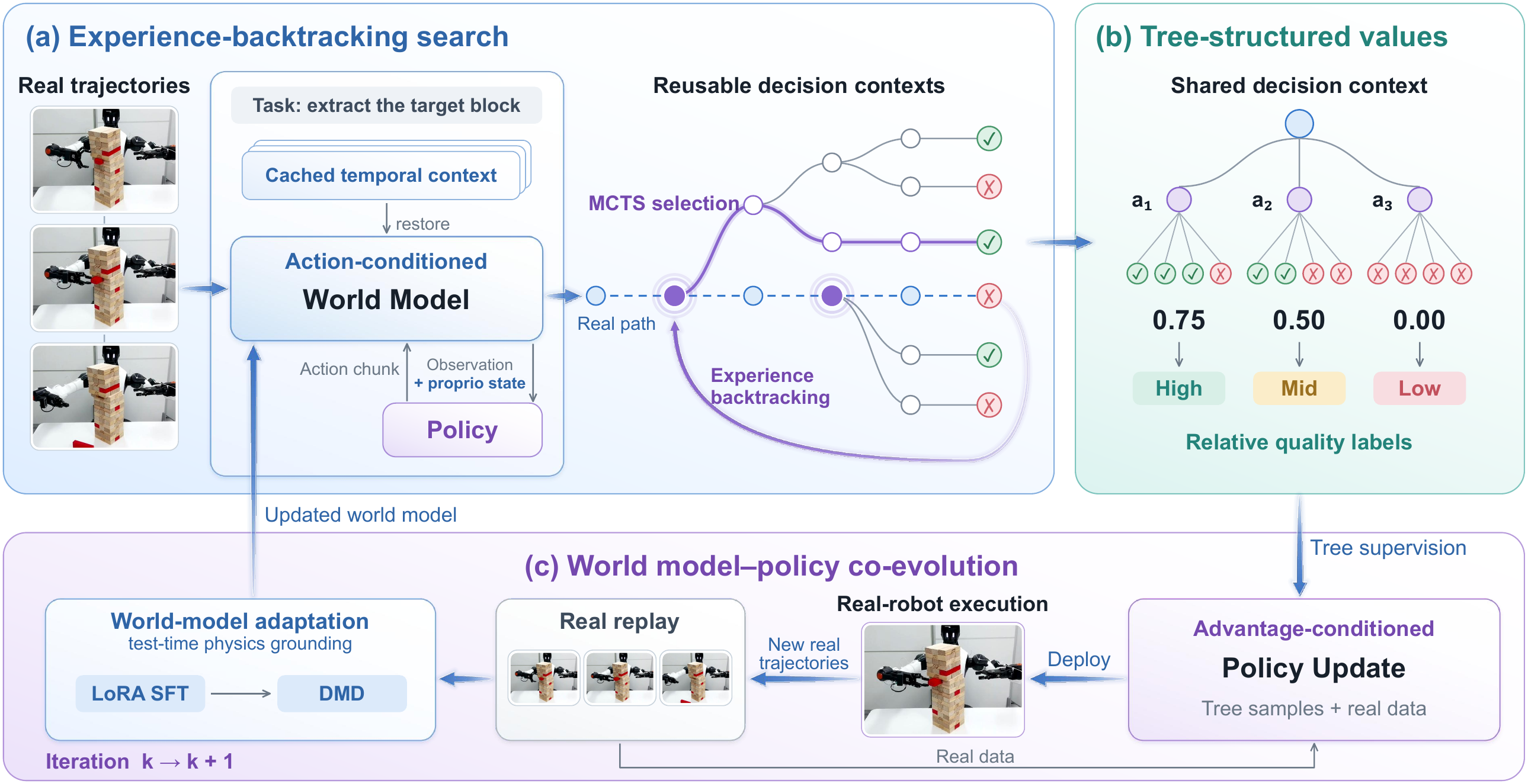}
    \captionsetup{format=loose}
\caption{
\textbf{Overview of \modelname{}.}
(a) Experience-backtracking tree search restores cached world-model contexts to explore informative branches without regenerating shared prefixes. The world model predicts both multi-view observations and the robot's proprioceptive state.
(b) Tree-structured value estimation aggregates downstream outcomes into relative action-quality labels for policy training without an additional value model.
(c) World model--policy co-evolution continually incorporates real trajectories to refine the world model through test-time physics grounding (LoRA SFT followed by DMD) while the policy keeps improving.
}
  \label{fig:framework}
\end{figure}

In this section, we elaborate on our proposed \modelname, a policy improvement paradigm that exploits experience backtracking in world models.
As illustrated in Figure~\ref{fig:framework}, \modelname revisits decision points, compares alternative actions through their downstream outcomes, and distills the results into the policy, while real-world feedback refines the world model.
\modelname follows the search-and-distill scheme of AlphaGo and Expert Iteration~\citep{alphago,expertiteration}, in which the policy proposes actions, search evaluates them, and the policy is then trained on the search results.
Unlike these methods, the search runs inside a learned video world model and starts from real robot experience instead of an exact simulator.
Search and training are also not alternating phases here: search, policy training, and world-model adaptation run concurrently and exchange data through a shared pool, and each tree uses the model versions that are current when it starts (Section~\ref{sec:system}).

\subsection{Experience-Backtracking Tree Search}
\label{sec:memory-search}

Independent rollouts expend substantial computation on repeatedly generating shared trajectory prefixes, yet provide limited comparisons of alternative actions from critical intermediate decision contexts.
World models enable these comparisons by restoring saved contexts and simulating alternative continuations from previously explored decision points.
Building on this capability, \modelname organizes reusable decision contexts into a search tree, allowing exploration to branch from selected nodes without regenerating their shared prefixes.

\paragraph{Decision Context Construction.}
\looseness=-1 Experience-backtracking search requires decision contexts for exploring alternative continuations.
Using GE-Sim 2.0 as the world model~\citep{qiu2026ge}, we construct a reference path by sampling a seed trajectory from offline demonstrations or collected real trajectories and retaining observations at action-chunk boundaries.
Given an action chunk, GE-Sim 2.0 jointly predicts the future frames of the three camera views and, through its pose-expert head, the robot's proprioceptive state (seven arm joints and one gripper per arm); together they form the imagined observation from which the policy proposes the next chunk.
Without the predicted proprioceptive state, the policy would have to infer the arm configuration from generated images alone, which is too coarse for the precise contacts and release timing that our tasks require.
Each non-terminal node serves as a branching point where the policy proposes alternative actions and the world model simulates their consequences from the corresponding real observation.
The outcome of the sampled trajectory is backed up along the reference path, so the real attempt itself acts as the first evaluated branch of the tree.
We cache the world-model context at each imagined node, enabling further continuations from the same decision point without regenerating the shared prefix.

\paragraph{MCTS-Guided Exploration.}
\looseness=-1 Building on these reusable decision contexts and inspired by AlphaGo~\citep{alphago}, we use Monte Carlo tree search (MCTS)~\citep{mcts} to concentrate a limited imagination budget on promising regions of the decision space.
By propagating rollout outcomes back to earlier decisions, MCTS uses evidence from completed explorations to guide subsequent branch selection. Branches with higher observed success rates receive further exploration, while an exploration bonus preserves opportunities for less-visited alternatives. This adaptive allocation helps discover successful continuations and examine alternative actions around them, producing informative comparisons for policy improvement rather than spending the budget uniformly across decision points.
For each node $v$, we maintain a visit count $N(v)$ and accumulated return $W(v)$, yielding the empirical search value $Q(v)=W(v)/N(v)$. An existing child $u$ is selected according to the PUCT score as follows:
\begin{equation}
    U(v,u)=Q(u)
    +c_{\mathrm{puct}}P(u\mid v)
    \frac{\sqrt{1+N(v)}}{1+N(u)},
    \label{eq:puct}
\end{equation}
where the first term exploits high empirical success rates and the second explores under-visited branches.
We assign uniform priors $P(u\mid v)$ over existing children.
For continuous action chunks, progressive widening~\citep{progressivewidening} permits at most
$\lceil c_{\mathrm{pw}}\max(1,N(v))^\alpha\rceil$ actions, where $0<\alpha<1$.
When permitted, a zero-valued virtual child with prior $1/(|\mathcal{C}(v)|+1)$, where $\mathcal{C}(v)$ is the set of existing children of $v$, competes under the same rule, balancing new actions against existing continuations.
Each new action is rolled out until success or the depth limit, after which a task-specific reward model returns $y\in\{0,1\}$.
We back up this return via $N\leftarrow N+1$ and $W\leftarrow W+y$.
For informative comparisons, we sample two actions at each new branch and prioritize rollouts in sibling subtrees with different provisional value bins.

\paragraph{Policy-Guided Expansion.}
Since action chunks are continuous and high-dimensional, the children of a node cannot be enumerated.
We use the policy itself as the proposal distribution: each expansion restores the context of the selected node and samples a chunk from the highest-bin policy $\pi_\theta(\cdot\mid o_v,g,b=K)$ with a fresh noise seed, where $o_v$ is the real or imagined observation at $v$ and $b=K$ is the highest of the $K$ value bins defined in Section~\ref{sec:tree-value}, so the search explores alternatives that the current policy considers plausible, in the manner of Sampled MuZero~\citep{sampledmuzero}.
The search stops once the number of simulated action chunks reaches the imagination budget $B$.

\paragraph{Disagreement-Driven Refinement.}
A pair of sibling actions provides a training signal only when their outcomes differ, and a difference observed from one or two rollouts can arise by chance.
Whenever sibling subtrees fall into different provisional value bins (Equation~\ref{eq:relative-bins}), we push their parent into a priority queue and allocate further rollouts to the least-visited sibling until every sibling has received at least $n_{\mathrm{conf}}$ visits.
This concentrates the budget on comparisons that produce supervision and reduces label noise before they are used for training; Algorithm~\ref{alg:rollback_search} in Appendix~\ref{app:training_algorithm} gives the complete procedure.

\subsection{Tree-Structured Value Estimation}
\label{sec:tree-value}
\looseness=-1
Tree structure connects an action to multiple downstream outcomes and allows alternative actions to be compared within a shared decision context. We use this structure to obtain chunk-level quality labels without fitting an additional value function.
Because labels are computed relative to sibling actions, decision contexts of different difficulty contribute comparable supervision.

\paragraph{Chunk-Level Value Estimation.}
\looseness=-1 We estimate the value of each action chunk from the outcomes of its downstream continuations. Each chunk $\mathbf{a}_i$ connects a parent node $v$ to an imagined child $u_i$, whose subtree contains the continuations explored after that action.
We obtain their outcomes using the world judge of GE-Sim 2.0~\citep{qiu2026ge}, a reward model that we fine-tune for each task on annotated robot videos with binary cross-entropy supervision.
Given visual observations and the task instruction, the model predicts task completion, which is converted into a binary outcome $y_\ell\in\{0,1\}$ using task-specific success criteria and temporal consistency checks. The chunk value is then estimated as follows:
\begin{equation}
    \hat{q}(v,\mathbf{a}_i)
    =\frac{1}{|\mathcal{L}(u_i)|}
    \sum_{\ell\in\mathcal{L}(u_i)}y_\ell,
    \label{eq:subtree-quality}
\end{equation}
where $\mathcal{L}(u_i)$ denotes the leaves in the subtree rooted at $u_i$.
This estimate measures the empirical success rate of the explored continuations without requiring an additional learned value critic. The reward model only judges whether a completed rollout succeeds, which is easier to train than a value model that predicts success from intermediate states.

\paragraph{Relative Value Supervision.}
To convert chunk-level value estimates into policy supervision, we compare alternative actions from the same decision context. Let $\mathcal{C}_{\mathrm{im}}(v)$ denote the imagined children of node $v$ and $q_{\max}(v)=\max_{u_i\in\mathcal{C}_{\mathrm{im}}(v)}\hat{q}(v,\mathbf{a}_i)$. For groups containing at least two actions with $q_{\max}(v)>0$, we normalize each chunk value by the best sibling value and discretize it into $K$ bins:
\begin{equation}
    \rho_i=\frac{\hat{q}(v,\mathbf{a}_i)}{q_{\max}(v)},
    \qquad
    b_i=\min\!\left(K,1+\lfloor K\rho_i\rfloor\right).
    \label{eq:relative-bins}
\end{equation}
\looseness=-1 The label $b_i\in\{1,\ldots,K\}$ indicates how well an action performs relative to the best explored alternative.
Normalizing within a group of siblings resembles the group-relative advantages of GRPO~\citep{grpo}, which compare samples generated for the same prompt; here, the group consists of actions sampled from the same restored context.
Groups whose siblings share one bin carry no preference signal and are discarded.
Each retained action contributes a sample $(o_v,g,\mathbf{a}_i,b_i)$ to $\mathcal{D}_{\mathrm{tree}}$ for policy training.

\paragraph{Advantage-Conditioned Policy Update.}
Following RECAP~\citep{recap}, we train the policy conditioned on discretized advantage labels, using our tree-derived relative value bins as chunk-level supervision. We instantiate the policy with $\pi_{0.5}$~\citep{pi05} and encode each bin in the policy prompt. For a normalized action chunk $\mathbf{a}$ and Gaussian noise $\boldsymbol{\epsilon}\sim\mathcal{N}(\mathbf{0},\mathbf{I})$, we construct $\mathbf{x}_s=(1-s)\mathbf{a}+s\boldsymbol{\epsilon}$ at flow time $s$ and optimize the flow-matching objective~\citep{flowmatching} as follows:
\begin{equation}
    \mathcal{L}_{\pi}(\theta)
    =
    \mathbb{E}_{\substack{
        (o,g,\mathbf{a},b)\sim\mathcal{D}_{\mathrm{mix}}\\
        s,\boldsymbol{\epsilon},\tilde b\sim p(\tilde b\mid b)
    }}
    \left[
        \left\|
        v_\theta(\mathbf{x}_s,s\mid o,g,\tilde b)
        -(\boldsymbol{\epsilon}-\mathbf{a})
        \right\|_2^2
    \right],
    \label{eq:policy-loss}
\end{equation}
where $\mathcal{D}_{\mathrm{mix}}$ combines tree-derived samples with offline demonstrations and robot-collected trajectories. Successful real trajectories receive the highest bin, and failed trajectories receive the lowest bin.
During training, we randomly omit the advantage condition with probability $p_{\mathrm{drop}}$. During tree search and deployment, we condition on the highest bin and apply classifier-free guidance~\citep{cfg} through $v_{\mathrm{guided}}=v_{\varnothing}+\beta(v_K-v_{\varnothing})$, where $\beta$ controls the guidance strength.
Because search proposals are sampled from the same highest-bin policy, each new policy snapshot also changes the actions proposed in the trees started after it.

\subsection{World Model--Policy Co-Evolution}
\label{sec:co-evolution}
\looseness=-1 As the policy evolves, its observation-action distribution may shift beyond the world model's training data, allowing policy optimization to exploit inaccurate predictions. We therefore keep adapting the world model through supervised fine-tuning (SFT) and distribution matching distillation (DMD) while the policy improves. For the $k$-th adaptation, the successful and failed trajectories collected since the previous one, $\mathcal{D}_{\mathrm{new}}^{(k)}$, are added to the replay set as \mbox{$\mathcal{D}_{\mathrm{real}}^{(k+1)}=\mathcal{D}_{\mathrm{real}}^{(k)}\cup\mathcal{D}_{\mathrm{new}}^{(k)}$}. We first adapt the world model to these real transitions using its action-conditioned prediction objective:
\begin{equation}
\mathcal{L}_{\mathrm{SFT}}^{(k)}(\phi)
=\mathbb{E}_{\tau\sim\mathcal{D}_{\mathrm{real}}^{(k+1)}}
\left[\sum_t
\ell_{\mathrm{WM}}\bigl(\phi;o_{\leq t},\mathbf{a}_t,o_{t+1},g\bigr)\right].
\label{eq:wm-sft}
\end{equation}
where $\ell_{\mathrm{WM}}$ is the action-conditioned prediction loss used to train the world model, applied to both the multi-view frames and the proprioceptive state.
We then apply distribution matching distillation (DMD)~\citep{dmd} to obtain a few-step generator for efficient imagined interaction. The resulting world model supports experience-backtracking search to generate new tree-structured supervision.
Because the newly collected trajectories are produced by recent policies, adaptation targets the states and actions that upcoming search is likely to visit.

\paragraph{Test-Time Physics Grounding.}
We call the capability that this adaptation provides test-time physics grounding.
Here test time refers to the world model, not the policy: it is the period after the world model's own pretraining in which it serves as the environment for policy learning.
The initial world model is fine-tuned on expert data, which covers the contacts and object motions that teleoperators produce but few of those produced by a policy that fails in new ways or succeeds along new routes; an error there, for example a block that stays upright when it should topple, can be exploited by search and labeled as success.
Grounding needs no extra annotation, since each real transition $(o_{\leq t},\mathbf{a}_t,o_{t+1})$ supplies its own target under Equation~\ref{eq:wm-sft}.
A LoRA adapter~\citep{lora} keeps each update small enough to repeat throughout training without retraining the full model, and DMD restores few-step sampling for search.
The capability is specific to learned simulators: a physics engine has hand-specified dynamics, and system identification can only tune its parameters.
It is also a general requirement for any learned simulator used to train policies, because an improving policy drifts toward the states that the simulator's training data covers least.
The frozen-world-model ablation in Section~\ref{sec:ablation} disables this grounding.

%% file: techreport/section/5_system.tex
\section{Asynchronous Real--Imagination Training System}
\label{sec:system}
\looseness=-1 Search in a video world model is computationally heavy, whereas real-robot interaction is slow and cannot be parallelized beyond the number of available robots.
A synchronous loop that alternates data collection, search, and training would leave each resource idle while the others work, so \modelname runs them as concurrent processes.
The three processes also run at different rates: a real episode lasts as long as the robot needs to complete or fail the task and be reset, a search tree lasts until it spends its imagination budget, which takes different amounts of time for different seeds, and the learners advance at the rate of their gradient steps.
None of them waits for another: they exchange only data and versioned model snapshots.
Section~\ref{sec:infra} describes where each process runs, Section~\ref{sec:search-coordination} how search is scheduled around cached world-model contexts, and Section~\ref{sec:training-process} the resulting training loop.

\begin{figure}[t]
  \centering
  \includegraphics[width=\linewidth]{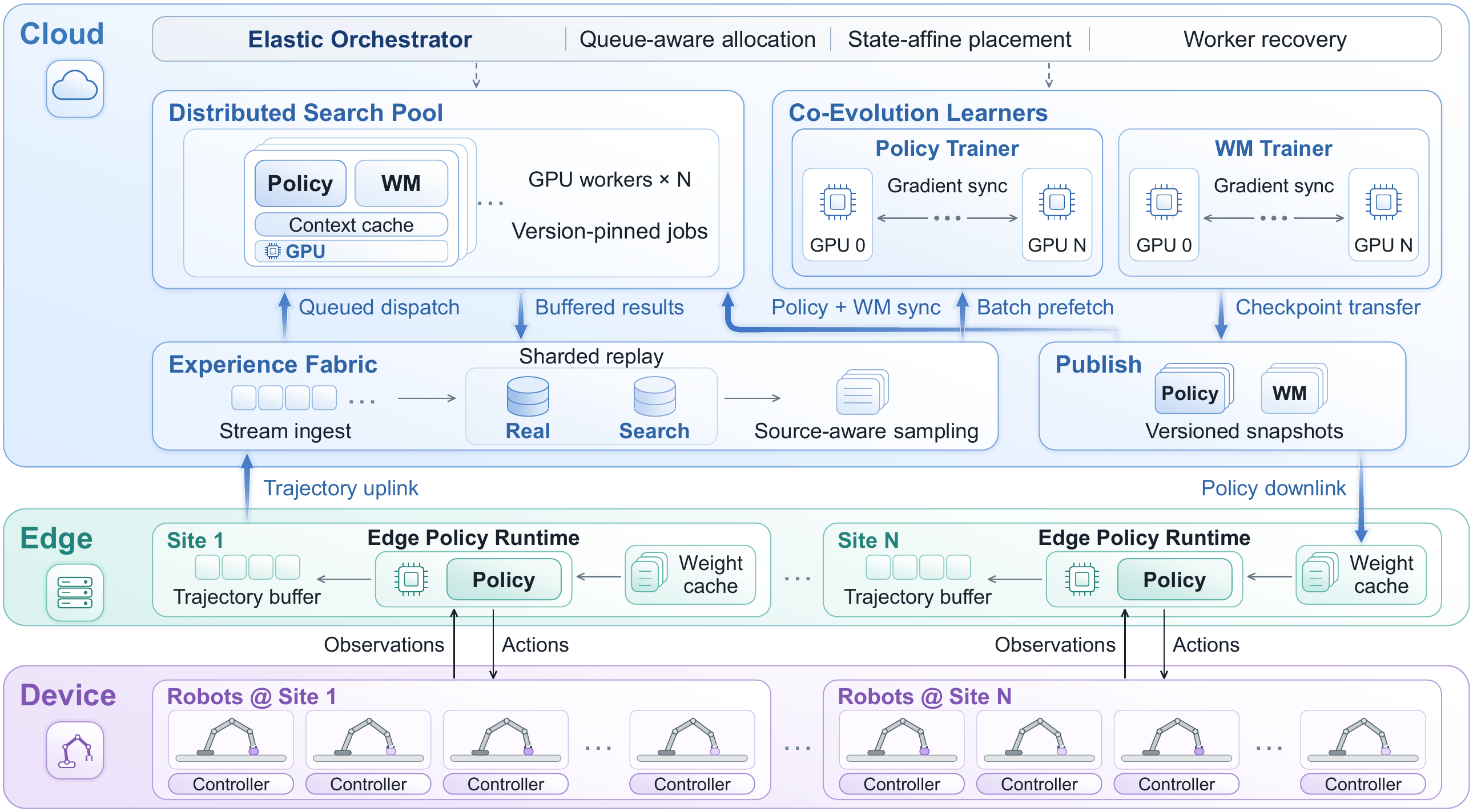}
  \caption{
  \textbf{\modelname training infrastructure.}
  The cloud runs tree search on GPU workers, stores real trajectories and tree supervision in the Experience Fabric, and trains the policy and the world model in parallel.
  Edge sites run policy inference for their robots, buffer trajectories for upload, and cache policy snapshots.
  Robots send observations to their edge site and execute the returned actions.
  Policy snapshots are sent to search workers and edge sites; world-model snapshots are sent only to search workers.
  }
  \label{fig:pipeline}
\end{figure}

\subsection{Cloud--Edge--Device Infrastructure}
\label{sec:infra}
Figure~\ref{fig:pipeline} divides the system into three tiers.
All GPU-intensive work, namely search, data storage, and training, runs in the cloud.
Edge sites are located next to the robots and run policy inference, so the cloud is not part of the robot's control loop.
Devices only sense and actuate.

\paragraph{Distributed Search Pool.}
Each search GPU hosts a policy replica, a world-model instance, and a cache of world-model contexts.
The Elastic Orchestrator assigns each worker seed trajectories drawn from offline demonstrations and collected real trajectories, and the worker builds the tree for a seed without communicating with other workers.
Policy sampling, world-model prediction, and context restoration all take place on the same GPU, so no intermediate rollout state is sent over the network during search; a worker sends only the supervision extracted from each finished tree.
Because search workers share no state, the throughput of imagination grows with the number of GPUs.

\paragraph{Experience Fabric.}
All data exchanged between producers and learners passes through the Experience Fabric.
It stores real trajectories and tree supervision in separate streams because they are used differently: offline demonstrations and real trajectories seed search and provide direct policy supervision, tree samples provide chunk-level quality labels, and world-model adaptation uses real trajectories only.
Policy batches are sampled from all three sources to form $\mathcal{D}_{\mathrm{mix}}$ and prefetched to the learners.
Robots and search workers write whenever data is ready, learners read whenever they need a batch, and no process waits for another; this shared pool is the only place where processes exchange data.

\paragraph{Co-Evolution Learners.}
\looseness=-1 The Policy Trainer optimizes Equation~\ref{eq:policy-loss} continuously with distributed data parallelism (DDP) on batches prefetched from the Experience Fabric.
Concurrently, the WM Trainer performs the test-time physics grounding described in Section~\ref{sec:co-evolution}: it repeatedly adapts the world model to real trajectories with LoRA SFT followed by DMD, using 400 and 100 steps per adaptation in our experiments.
Gradients are synchronized only within each trainer, and neither trainer waits for the other or for search.
Both trainers hand their weights to the Publish component, which stores them as versioned snapshots and sends policy snapshots to search workers and edge sites, and world-model snapshots to search workers.

\paragraph{Edge Sites and Devices.}
Each edge site runs a Policy Runtime, a trajectory buffer, and a weight cache.
A robot streams observations to the Policy Runtime and executes the returned action chunks with its own low-level controller.
New policy snapshots are downloaded into the weight cache, and the runtime switches a robot to the new version only when its current episode ends, so each real trajectory is generated by a single policy.
Robots at the same site switch independently.
The trajectory buffer turns each episode's observations, actions, and outcome into a labeled trajectory and uploads it to the Experience Fabric in the background.

In our experiments, the cloud consists of 48 NVIDIA H200 GPUs, of which 40 run world-model search and 8 run policy training, and serves one edge site with one physical robot.
Additional sites would connect through the same upload and download paths; we have not evaluated such configurations.

\subsection{Search Scheduling and Context Reuse}
\label{sec:search-coordination}

\paragraph{Worker-Local Contexts.}
Experience backtracking branches from states that earlier rollouts have already reached.
Regenerating the prefix of each new branch would cost as many world-model calls as the depth of the branching node.
Each worker therefore keeps its trees and the world-model contexts of their nodes in GPU memory, and a new branch starts by restoring the cached context of the selected node (Section~\ref{sec:memory-search}).
A tree stays on the worker that holds its contexts until it is finished, and the orchestrator sends new seeds to idle workers.

\paragraph{Version-Pinned Search Jobs.}
Trees take different amounts of time because their seed trajectories, depths, and refinement steps differ.
If search paused for every model update, each update would wait for the slowest tree, and the search GPUs would sit idle while the learners train.
Search and training are therefore fully decoupled: workers keep searching while both trainers keep training.
Each search job is pinned to the latest policy and world-model snapshots available when its tree starts and keeps them until the tree is finished; snapshots published in the meantime are picked up by the worker's next tree.
Pinning keeps every tree internally consistent: all of its branches are proposed by one policy and simulated by one world model, so sibling outcomes remain comparable and the cached contexts stay valid for the whole tree.
Supervision from a finished tree enters the Experience Fabric immediately and is mixed into subsequent policy batches.

\paragraph{Cost of Imagination.}
Two mechanisms reduce the GPU time spent on each tree.
Context reuse removes the recomputation of shared prefixes, and DMD reduces the number of sampling steps needed to generate each new action chunk (Section~\ref{sec:co-evolution}).
Network traffic is limited to seed dispatch, finished supervision, and weight synchronization, since all tree operations stay on the worker.
Search capacity can therefore be increased by adding GPUs, whereas real data grows only with the number of robots.
As search GPUs are added, the Policy Trainer and the Experience Fabric must process proportionally more tree supervision, so they eventually limit throughput.

\subsection{Training Process}
\label{sec:training-process}

\begin{figure}[t]
  \centering
  \includegraphics[width=\linewidth]{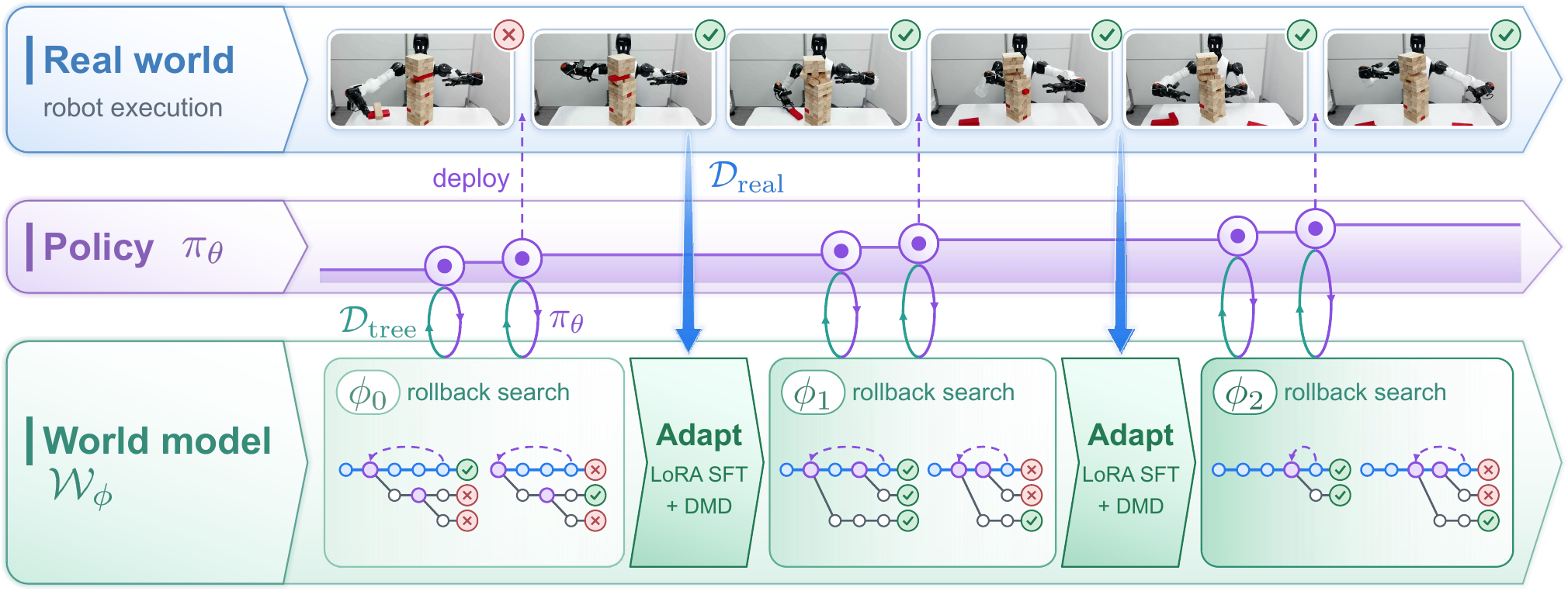}
\caption{
\textbf{Training process of \modelname.} The robot continually executes the latest policy and uploads real trajectories $\mathcal{D}_{\mathrm{real}}$. The policy is continually updated with tree supervision $\mathcal{D}_{\mathrm{tree}}$ from rollback search, while the world model is repeatedly adapted to real trajectories via LoRA SFT and DMD; each search tree uses the latest versions of both.
}
  \label{fig:training-process}
\end{figure}

As shown in Figure~\ref{fig:training-process}, \modelname runs real-robot execution, world-model search, policy training, and world-model adaptation concurrently, exchanging information only through the Experience Fabric and asynchronous transfer of versioned snapshots.
The robot continually executes the latest policy it has received and uploads each labeled trajectory to $\mathcal{D}_{\mathrm{real}}$; new weights take effect only between episodes, so physical data collection never waits for training.
Meanwhile, GPU workers continually run experience-backtracking search from offline demonstrations and the growing $\mathcal{D}_{\mathrm{real}}$, and the Policy Trainer continually updates the policy on $\mathcal{D}_{\mathrm{mix}}$ with Equation~\ref{eq:policy-loss}, so tree supervision is used as soon as it is produced.
In parallel, the WM Trainer repeatedly adapts the world model to the accumulated real trajectories (Section~\ref{sec:co-evolution}) and publishes each adapted model to the search pool.
A policy snapshot thus reaches the robot at its next episode boundary and each search worker at its next tree, and the trajectories it collects reach the world model at the next adaptation.
Algorithm~\ref{alg:rollback_training} gives the complete training procedure.

%% file: techreport/section/6_experiment.tex
\section{Experiments}
\label{sec:experiments}
\looseness=-1 We evaluate \modelname{} on three real-world manipulation tasks to assess policy performance and data efficiency. We investigate the contributions of experience-backtracking search (through the comparison with RISE w/ GE-Sim2, our reproduction of RISE on the same world model without backtracking), tree-structured supervision, and world-model adaptation to iterative policy improvement.
Specifically, we ask whether \modelname outperforms existing baselines in the real world, how much each component contributes, whether it improves stably and reaches human-level performance, and how faithful its world model is.

\subsection{Experimental Setup}
\label{sec:setup}

\paragraph{Tasks and Evaluation.}
We evaluate our proposed \modelname{} and baselines on the AGIBOT Genie-02 robot~\citep{agibotg2} across three real-world tasks spanning contact-rich manipulation, precise control, and competitive interaction:
\begin{itemize}[leftmargin=*]
    \item \textit{Block Extraction.}
    The robot removes wooden Jenga blocks one at a time from a stacked tower without collapsing it, testing precise manipulation under stability constraints.
    \item \textit{Ball Tossing.}
    The robot throws a ball into a bucket, with success determined by whether the ball lands inside. This task tests precise control of the throwing motion and release timing.
    \item \textit{Tabletop Curling.}
    \looseness=-1 A human and the robot each deliver two stones. Later shots can displace earlier stones, and the winner is determined by the score based on proximity to the target center. This task tests strategic selection and precise execution against an opponent.
\end{itemize}
The policy observes three RGB views, from one head-mounted camera and two wrist cameras, together with the robot's proprioceptive state, and controls 16 action dimensions, comprising seven arm joints and one gripper per side, at 30\,Hz.
For each task, we select six initial positions (nine for Tabletop Curling) and run 15 trials per position (25 for Block Extraction); per-position results are in Appendix~\ref{app:per-position}.

\paragraph{Implementation Details.}
\textit{(1) World Model Training.}
We initialize the world model from pretrained GE-Sim 2.0~\citep{qiu2026ge} and train it in two stages: supervised fine-tuning followed by distribution matching distillation~\citep{dmd}. We fine-tune the model on task-specific data for 10K steps using 32 NVIDIA H200 GPUs with a global batch size of 128. We then perform DMD on the same dataset for 2,500 steps using 16 NVIDIA H200 GPUs with a global batch size of 16.
\textit{(2) Policy Warm-up.}
We initialize the policy from $\pi_{0.5}$~\citep{pi05} and pre-train it on task-specific expert-collected data using RECAP-style quality conditioning~\citep{recap}. We use five quality bins, assigning successful trajectories to the highest bin $5$ and failed trajectories to the lowest bin $1$. Training runs for 100K steps on 16 NVIDIA H200 GPUs with a global batch size of 64.
\textit{(3) Reward Model Training.}
We use the world judge of GE-Sim 2.0~\citep{qiu2026ge} as the reward model and fine-tune it on task-specific robot videos annotated with success labels. Conditioned on visual observations and the task instruction, the model predicts task completion using a binary classification head trained with binary cross-entropy loss.
\textit{(4) Policy Improvement in World Models.}
We run policy improvement with one physical robot and 48 NVIDIA H200 GPUs, allocating 40 GPUs to world-model-based tree search and 8 to policy training with DDP. Following world-model warm-up, all processes run asynchronously; each world-model adaptation adds newly collected real trajectories to the replay dataset and adapts the world model through 400 steps of LoRA-based~\citep{lora} SFT followed by 100 steps of DMD.

\paragraph{Baselines.}
\looseness=-1 We compare \modelname{} with imitation learning, reinforcement learning, and world-model-based baselines. All methods use the same $\pi_{0.5}$ initialization and comparable compute budgets; RECAP and RISE w/ GE-Sim2 are trained for a similar duration of about 8 hours with one physical robot and a similar number of NVIDIA H200 GPUs. Execution-level details and hyperparameters are provided in Appendices~\ref{app:details} and~\ref{app:hparams}.
\textit{(1) SFT.} We fine-tune $\pi_{0.5}$~\citep{pi05} on task-specific expert demonstrations using supervised learning.
\textit{(2) RECAP.} We adapt RECAP~\citep{recap} to $\pi_{0.5}$, using a learned value model to estimate advantages for conditioned policy training on collected real-world data.
\textit{(3) RISE w/ GE-Sim2.} RISE~\citep{rise} learns from imagined interactions with a compositional world model that couples action-conditioned dynamics with a learned progress value model.
We reproduce RISE on GE-Sim 2.0~\citep{qiu2026ge}, the same world model that \modelname{} uses, with five RECAP-style advantage bins, and denote this reproduction RISE w/ GE-Sim2. All RISE results in this report come from this reproduction rather than from the original paper. Because the two methods share the world model, the comparison isolates experience-backtracking search and tree-structured supervision from world-model quality.

\subsection{Main Results}
\looseness=-1 Table~\ref{tab:main-results} compares \modelname with the baselines across three real-world tasks. \modelname{} achieves the highest success rate on all tasks, reaching 69.33\% on Block Extraction, 40.00\% on Ball Tossing, and 64.44\% on Tabletop Curling. \modelname's average success rate exceeds SFT, RECAP, and RISE w/ GE-Sim2 by 17.63\%, 13.71\%, and 11.21\%, respectively. Compared with RISE w/ GE-Sim2, which also learns through interaction with the same world model, \modelname improves success rates by an absolute 10.67\%, 11.11\%, and 11.85\% on the three tasks. This corresponds to a relative improvement of 24.0\% in average success rate over RISE w/ GE-Sim2.
On average, real-world RL with RECAP adds 3.92\% over SFT and forward world-model rollouts with RISE w/ GE-Sim2 add another 2.49\%, whereas experience backtracking with the same world model adds 11.21\% on top of RISE w/ GE-Sim2, more than the two previous gains combined.
The largest relative gain is on Ball Tossing (38.5\% over RISE w/ GE-Sim2), where the outcome is decided by the release, so comparing sibling actions from the same context directly identifies better releases.
These gains span contact-rich manipulation, precise throwing, and competitive interaction, indicating that the benefits of search-based training transfer to physical execution without requiring world-model search at deployment.

\input{techreport/table/main}

\input{techreport/table/ablation}

\subsection{Ablation Studies}
\label{sec:ablation}
We conduct ablations on Block Extraction to examine the contributions of world-model adaptation and tree-structured value estimation; frozen-world-model results on the other two tasks are reported in Appendix~\ref{app:per-position}.
We report three metrics over all evaluation trials: \textit{Target Contact} measures whether the pushing motion hits the target block; \textit{Task Success} measures complete extraction without collapsing the remaining structure; and \textit{Consecutive Extractions} counts the average number of blocks extracted in a row before failure.
\textit{(1) Frozen World Model.}
We retain the world model obtained after initial task-specific training and disable subsequent updates from real interaction, which disables test-time physics grounding.
\textit{(2) Direct Reward Supervision.}
We replace subtree-aggregated values with individual rollout rewards for quality-conditioned policy training while retaining tree search.
All variants use the same policy initialization and computation budget.
As shown in Table~\ref{tab:ablation}, removing either component degrades all three metrics; direct reward supervision lowers task success most (59.33\%), and a frozen world model reduces consecutive extractions from 2.60 to 1.90.
Target contact also falls from 89.33\% to 78.67\% with a frozen world model and to 77.33\% with direct reward supervision.
With direct reward supervision, each label depends on a single stochastic rollout rather than a subtree average, and with a frozen world model, prediction errors accumulate over the long sequences of contact-rich decisions that block extraction requires.

\begin{figure}[t]
  \centering
  \begin{minipage}[t]{0.485\linewidth}
    \centering
    \includegraphics[width=\linewidth]{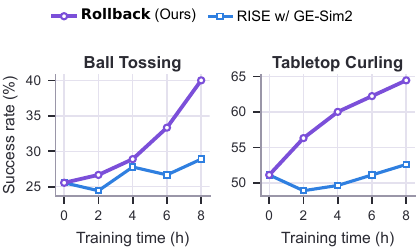}
    \caption{
    \textbf{Training dynamics.} Real-world success rate of \modelname{} and RISE w/ GE-Sim2 versus wall-clock training time.
    }
    \label{fig:training-curves}
  \end{minipage}\hfill
  \begin{minipage}[t]{0.485\linewidth}
    \centering
    \includegraphics[width=\linewidth]{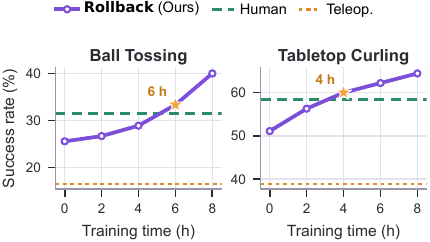}
        \caption{
    \textbf{Surpassing human performance.} Success rate versus training time against human players and teleoperation (dashed). Stars mark when \modelname{} first exceeds humans.
    }
    \label{fig:human}
  \end{minipage}
\end{figure}

\subsection{Training Dynamics}
\label{sec:training-analysis}
Figure~\ref{fig:training-curves} tracks the real-world success rate of \modelname{} and RISE w/ GE-Sim2 on Ball Tossing and Tabletop Curling over 8 hours of training. Starting from the RECAP policy, \modelname{} improves Ball Tossing steadily from 25.56\% to 40.00\%, whereas RISE w/ GE-Sim2 fluctuates and ends at only 28.89\%.
This is a relative improvement of 56.5\% within 8 hours using a single physical robot, while RISE w/ GE-Sim2 first drops to 24.44\% after 2 hours. Keeping the world model frozen instead yields only 26.67\%, indicating that continually refining the world model with real experience is essential for sustained improvement.
On Tabletop Curling, \modelname{} rises from 51.11\% to 64.44\% over the same period, while RISE w/ GE-Sim2 stays close to its initialization and ends at 52.59\%, only 1.48\% above its starting point.
On Tabletop Curling, the gap between the two methods widens at every evaluation; on Ball Tossing, it widens steadily after 4 hours.

\begin{figure}[p]
  \centering
  \includegraphics[width=\linewidth]{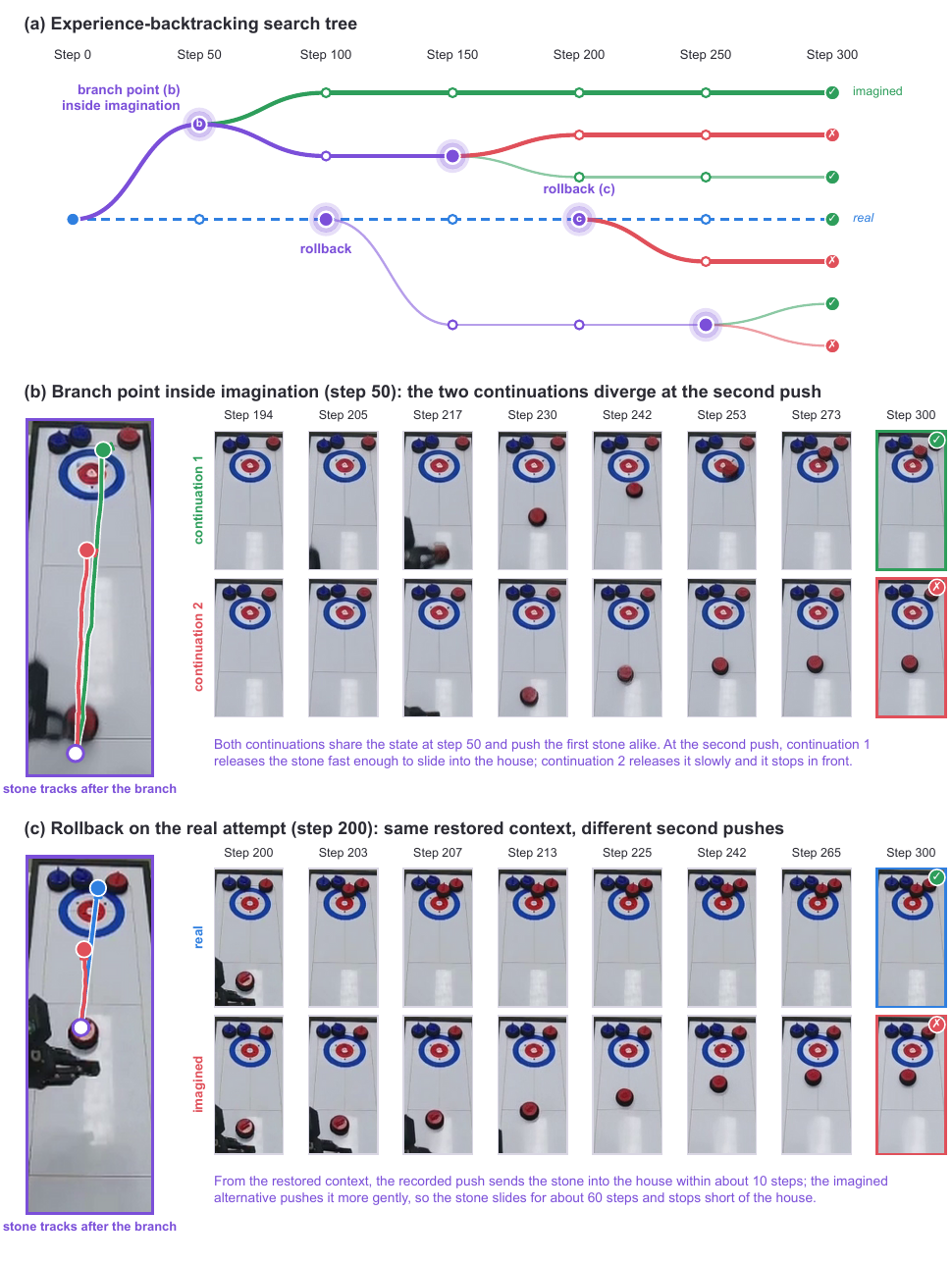}
\caption{
\textbf{Experience-backtracking search on Tabletop Curling.} (a) Search tree. The blue dashed path is a real attempt and colored paths are imagined continuations; edges are green, red, or purple when the continuations below all succeed, all fail, or are mixed. Purple rings mark branch points at intermediate decisions, each followed by successful and failed continuations. (b) Two world-model continuations from a branch point inside imagination; (c) the real attempt and a world-model alternative from a rollback point. Each panel shows stone tracks overlaid on the frame at the branch (left) and time-aligned frames (right). In both cases the continuations differ in how fast the second stone is released: a fast release reaches the house, a slow one stops short.
}
  \label{fig:qualitative}
\end{figure}

\subsection{Qualitative Results}
\label{sec:qualitative}
\looseness=-1 Figure~\ref{fig:qualitative} visualizes experience-backtracking search on Tabletop Curling: panel (a) shows the search tree, and panels (b, c) show continuations from a branch point inside imagination and from a rollback on a real attempt. Rather than restarting from the initial state, \modelname rolls back to two intermediate decision points of a real attempt and branches from the restored contexts. At step 200 (control steps at 30\,Hz), during the second push, the recorded push brings the red stone into the house, whereas an imagined alternative from the same context leaves it short of the house. Comparing continuations that share everything up to this point isolates the second push as the decision that determines the outcome. Branching also happens inside imagination: from the same imagined state at step 50, both continuations push the first stone alike, but one releases the second stone fast enough to reach the house and the other releases it slowly, so it stops in front. More examples for all three tasks are provided in Appendix~\ref{app:qualitative}.

\subsection{Surpassing Human Performance}
\label{sec:human}
We compare \modelname{} with human players performing each task by hand and with teleoperated data collection, both under the same initial positions and success criteria; each is averaged over three people.
On Ball Tossing, human players and teleoperation achieve 31.48\% and 16.53\%, respectively, and \modelname{} surpasses the human average after 6 hours of training (33.33\%), reaching 40.00\% after 8 hours (Figure~\ref{fig:human}).
On Tabletop Curling, human players and teleoperation achieve 58.39\% and 38.94\%, and \modelname{} surpasses humans after 4 hours, reaching 64.44\% after 8 hours.
Although \modelname{} is initialized from teleoperated demonstrations far below human level, it surpasses human players on both tasks by learning from counterfactual imagination.
At the end of training, \modelname{} exceeds the human average by 8.52\% and 6.05\% on the two tasks, and teleoperation by 23.47\% and 25.50\%.
The RECAP initialization already exceeds teleoperation on both tasks (25.56\% vs.\ 16.53\% and 51.11\% vs.\ 38.94\%), but only training with \modelname lifts it above human players.
The policy never observes the human players, so the margin over them comes from comparing the robot's own alternative actions in imagination.

\input{techreport/table/fidelity}
\input{techreport/table/fidelity_coevolution}
\begin{figure}[t]
  \centering
  \includegraphics[width=\linewidth]{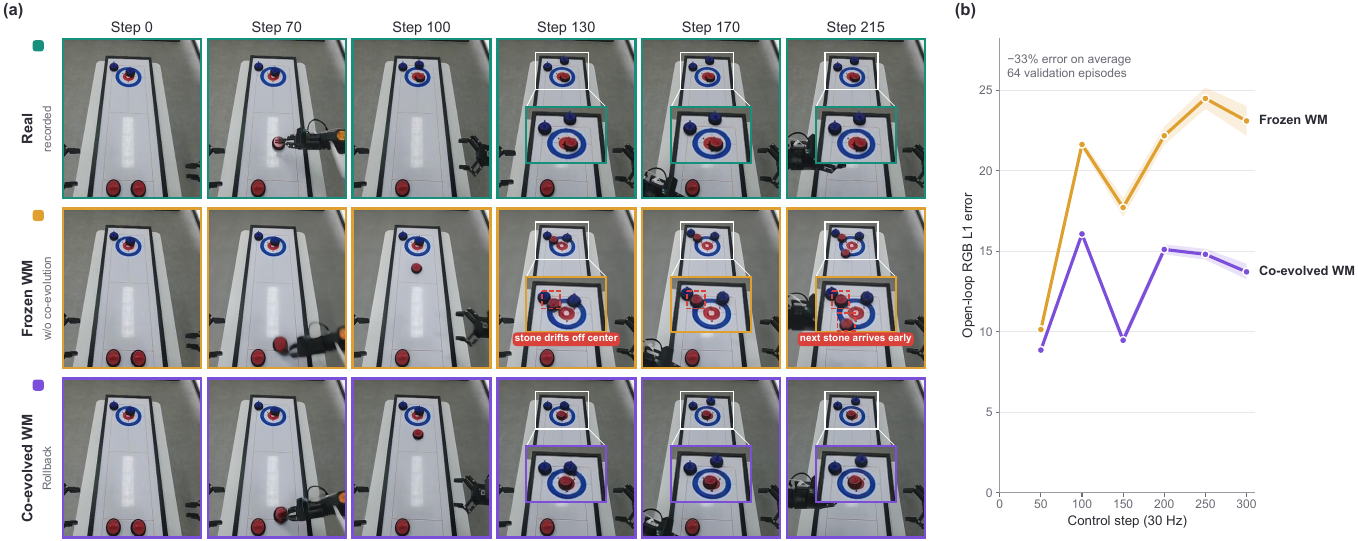}
\caption{
\textbf{World-model fidelity with and without co-evolution on Tabletop Curling.} (a) Open-loop predictions of the frozen world model (before co-evolution) and the co-evolved world model under the recorded actions of one validation episode, with the real recording on top. Each cell shows the head camera cropped to the sheet; insets magnify the house, and red boxes mark hallucinated stones. Steps are control steps at 30\,Hz. (b) RGB L1 error over control steps, averaged over the 64 validation episodes, with 95\% confidence intervals shaded.
}
  \label{fig:app-fidelity}
\end{figure}
\subsection{World-Model Fidelity and Reward Reliability}
\label{sec:fidelity}
To measure how far imagined rollouts drift from reality, we replay the recorded actions of eight failed real Block Extraction episodes open-loop in the co-evolved world model, five times each, and compare the predictions with the real recordings (Table~\ref{tab:app-fidelity}).
The average PSNR is 21.6\,dB, highest for the head camera (23.4\,dB), whose static background is easier to predict than the wrist views.
The per-pixel L1 error grows from 6.8 in the first quarter of each rollout to 11.8 in the last quarter, a 1.74$\times$ increase over rollouts of 150--275 frames, consistent with our choice to branch from real observations rather than roll out long imagined horizons.
Figure~\ref{fig:app-jenga-fidelity} in Appendix~\ref{app:qualitative} shows one of these episodes.
The reward model reports success on none of the 40 rollouts, all of which replay failed episodes.
A false positive would turn a failed imagined branch into a positive label and thereby corrupt the relative value labels that the policy is trained on.

To measure what test-time physics grounding changes, we replay the recorded actions of the 64 Tabletop Curling validation episodes open-loop, three times each, in the world model before co-evolution (frozen) and in the world model after co-evolution (co-evolved), and compare both with the real recordings (Table~\ref{tab:fidelity-coevolution}). These episodes are real trajectories collected by the policy, so the comparison measures fidelity on the states that the policy visits; they are part of the real data the co-evolved world model is adapted to and are not seen by the frozen one.
The co-evolved world model has a lower RGB L1 error on all 64 episodes, and its average error falls from 19.52 to 12.96.
The gap widens along the rollout: the error rises from 12.48 in the first quarter to 23.73 in the last quarter for the frozen world model (1.90$\times$), but only from 10.50 to 14.47 for the co-evolved one (1.38$\times$), as shown in Figure~\ref{fig:app-fidelity}(b).
The three rollouts of the co-evolved world model also differ less from each other (L1 spread 6.12 vs.\ 9.39).

Figure~\ref{fig:app-fidelity}(a) shows one episode. After the first stone stops at the center of the house, the frozen world model moves it to the edge of the house by step 130 and keeps it there, and at step 215 it brings the second stone into the house while the real stone is still sliding. The co-evolved world model keeps the first stone at the center and the second stone on the sheet, as in the recording. Such errors change the score of an imagined shot and thus the value labels of the branches that contain it, which the policy is then trained on.

%% file: techreport/table/main.tex
\begin{table}[t]
    \centering
    \caption{
\textbf{Performance on three real-world tasks.}
Success rates (\%) of \modelname{} and imitation-learning, reinforcement-learning, and world-model-based baselines.
    }
    \label{tab:main-results}
    \renewcommand{\arraystretch}{1.15}
    \small
    \setlength{\tabcolsep}{4pt}
    \begin{tabular*}{\linewidth}{@{\extracolsep{\fill}}llcccc}
        \toprule
        Method & Category & Block Extraction & Ball Tossing & Tabletop Curling & Average \\
        \midrule
        $\pi_{0.5}$ + SFT & Imitation learning
        & 48.67 & 23.33 & 48.89 & 40.30 \\
        $\pi_{0.5}$ + RECAP & Real-world RL
        & 56.00 & 25.56 & 51.11 & 44.22 \\
        $\pi_{0.5}$ + RISE w/ GE-Sim2 & World-model RL
        & 58.67 & 28.89 & 52.59 & 46.72 \\
        \midrule
        \rowcolor{niceblue!8}
        \textbf{\modelname{} (Ours)} & World-model RL
        & \textbf{69.33} & \textbf{40.00} & \textbf{64.44} & \textbf{57.93} \\
        \textit{Gain over strongest baseline} &
        & \textcolor{rbgain}{\textbf{+10.67}} & \textcolor{rbgain}{\textbf{+11.11}} & \textcolor{rbgain}{\textbf{+11.85}} & \textcolor{rbgain}{\textbf{+11.21}} \\
        \bottomrule
    \end{tabular*}
\end{table}

%% file: techreport/table/ablation.tex
\begin{table}[t]
    \centering
    \captionsetup{format=loose}
    \caption{
    \textbf{Ablation study on Block Extraction.}
    We evaluate world-model adaptation and tree-structured value estimation via target contact rate, task success rate, and the average number of consecutive extractions; the full framework performs best on all metrics.
    }
    \label{tab:ablation}
    \renewcommand{\arraystretch}{1.15}
    \small
    \setlength{\tabcolsep}{14pt}
    \begin{tabular}{lccc}
        \toprule
        Variant & Target Contact (\%) & Task Success (\%) & Consecutive Extractions \\
        \midrule
        Frozen World Model
        & 78.67 & 64.67 & 1.90 \\
        Direct Reward Supervision
        & 77.33 & 59.33 & 1.73 \\
        \midrule
        \rowcolor{niceblue!8}
        \textbf{\modelname{} (Full)}
        & \textbf{89.33} & \textbf{69.33} & \textbf{2.60} \\
        \bottomrule
    \end{tabular}
\end{table}

%% file: techreport/table/fidelity.tex
\begin{table}[t]
    \centering
    \caption{\textbf{Open-loop world-model fidelity on Block Extraction} (40 rollouts).}
    \label{tab:app-fidelity}
    \small
    \setlength{\tabcolsep}{6pt}
    \begin{tabular}{cccccc}
        \toprule
        \multicolumn{3}{c}{PSNR (dB) $\uparrow$} & \multicolumn{2}{c}{RGB L1 $\downarrow$} & \multirow{2}{*}{Reward false positives} \\
        \cmidrule(lr){1-3}\cmidrule(lr){4-5}
        Head & Left wrist & Right wrist & First quarter & Last quarter & \\
        \midrule
        23.41 & 20.63 & 20.85 & 6.79 & 11.79 & 0 / 40 \\
        \bottomrule
    \end{tabular}
\end{table}

%% file: techreport/table/fidelity_coevolution.tex
\begin{table}[t]
    \centering
    \caption{\textbf{Open-loop world-model fidelity with and without co-evolution on Tabletop Curling} (64 validation episodes, three rollouts each).}
    \label{tab:fidelity-coevolution}
    \small
    \setlength{\tabcolsep}{5pt}
    \begin{tabular}{lcccccc}
        \toprule
        \multirow{2}{*}{World model} & \multicolumn{3}{c}{PSNR (dB) $\uparrow$} & \multicolumn{2}{c}{RGB L1 $\downarrow$} & \multirow{2}{*}{Rollout spread $\downarrow$} \\
        \cmidrule(lr){2-4}\cmidrule(lr){5-6}
         & Head & Left wrist & Right wrist & First quarter & Last quarter & \\
        \midrule
        Frozen & 21.50 & 17.07 & 15.01 & 12.48 & 23.73 & 9.39 \\
        Co-evolved & \textbf{22.27} & \textbf{17.79} & \textbf{16.37} & \textbf{10.50} & \textbf{14.47} & \textbf{6.12} \\
        \bottomrule
    \end{tabular}
\end{table}

%% file: techreport/section/7_conclusion.tex
\FloatBarrier
\section{Conclusion}
\label{sec:conclusion}
We presented \modelname{}, a policy improvement paradigm that exploits experience backtracking in world models.
By revisiting decision contexts and aggregating the outcomes of alternative continuations, \modelname{} concentrates imagination on informative comparisons and obtains chunk-level supervision without a learned value model.
A unified training system couples this search with real-robot execution, jointly improving the policy and the world model through real-world feedback.
This test-time physics grounding keeps the world model accurate in the states the policy visits.
Experiments on three challenging real-world tasks demonstrate higher success rates and more stable improvement than competing baselines.
Over RISE w/ GE-Sim2, our reproduction of RISE on the same world model, \modelname improves the success rate by an absolute 10.67\%, 11.11\%, and 11.85\% on block extraction, ball tossing, and tabletop curling, and it surpasses the average human player on the latter two tasks after 8 hours of training with one robot.

%% file: techreport/section/8_appendix.tex
\clearpage
\appendix
\raggedbottom
\setcounter{algorithm}{0}
\renewcommand{\topfraction}{0.9}
\renewcommand{\bottomfraction}{0.8}
\renewcommand{\textfraction}{0.07}
\renewcommand{\floatpagefraction}{0.8}

\section{Detailed Training Algorithm}
\label{app:training_algorithm}

Algorithm~\ref{alg:rollback_training} summarizes the overall training procedure.
Real-robot execution, tree search, policy training, and world-model adaptation
run as concurrent processes that never wait for one another.  They exchange data
only through the shared pool of the Experience Fabric ($\mathcal D_{\mathrm{real}}$
and $\mathcal B_{\mathrm{tree}}$) and models only as versioned snapshots; each
search tree is pinned to the snapshots that are current when it starts.
Algorithm~\ref{alg:rollback_search} details the experience-backtracking
search, and Algorithm~\ref{alg:rollback_labels} gives the exact rule used to extract
tree-structured supervision.

\begin{algorithm}[H]
\caption{\modelname{} training with world model--policy co-evolution}
\label{alg:rollback_training}
\small
\begin{algorithmic}[1]
\Require Offline demonstrations $\mathcal D_{\mathrm{demo}}$; initial real replay
buffer $\mathcal D_{\mathrm{real}}$; policy $\pi_\theta$; world model
$\mathcal W_\phi$; task-specific GE-Sim 2.0 world judge; number of value bins $K$
\Ensure Updated policy $\pi_\theta$ and world model $\mathcal W_\phi$
\State Initialize an empty tree buffer $\mathcal B_{\mathrm{tree}}$ in the shared pool
\State Run the following four processes concurrently until training ends
\Statex \textbf{Real robot}
\Loop
    \State Load the latest policy snapshot at the episode boundary
    \State Execute one episode and append the labeled trajectory to $\mathcal D_{\mathrm{real}}$
\EndLoop
\Statex \textbf{Search worker} (one per search GPU)
\Loop
    \State Pin the latest snapshots $\pi_\theta$ and $\mathcal W_\phi$ for this tree
    \State Sample a seed trajectory
    $\tau\sim\mathcal D_{\mathrm{demo}}\cup\mathcal D_{\mathrm{real}}$
    \State $\mathcal T_\tau\gets
    \Call{BacktrackingTreeSearch}{\tau,\pi_\theta,\mathcal W_\phi}$
    \State Append $\Call{ExtractTreeLabels}{\mathcal T_\tau,K}$ to $\mathcal B_{\mathrm{tree}}$
\EndLoop
\Statex \textbf{Policy Trainer}
\Loop
    \State Sample a batch of $\mathcal D_{\mathrm{mix}}$ from
    $\mathcal B_{\mathrm{tree}}$, $\mathcal D_{\mathrm{demo}}$, and
    $\mathcal D_{\mathrm{real}}$
    \State Update $\theta$ with the advantage-conditioned policy objective (DDP)
    \State Periodically publish $\pi_\theta$ to search workers and the edge policy runtime
\EndLoop
\Statex \textbf{WM Trainer}
\Loop
    \State Adapt $\mathcal W_\phi$ on the current $\mathcal D_{\mathrm{real}}$ by LoRA SFT using
    the action-conditioned prediction objective
    \State Distill the adapted model into the few-step simulator by DMD
    \State Publish $\mathcal W_\phi$ to search workers
\EndLoop
\State \Return $\pi_\theta,\mathcal W_\phi$
\end{algorithmic}
\end{algorithm}

\begin{algorithm}[!t]
\caption{Experience-backtracking tree search}
\label{alg:rollback_search}
\small
\begin{algorithmic}[1]
\Require Seed trajectory $\tau$ with instruction $g$ and binary outcome
$y_\tau\in\{0,1\}$; policy $\pi_\theta$; world model $\mathcal W_\phi$;
task-specific GE-Sim 2.0 world judge; imagination budget $B$; maximum search depth
$D_{\max}$; number of bins $K$; PUCT coefficient
$c_{\mathrm{puct}}$; widening parameters $(c_{\mathrm{pw}},\alpha)$;
confirmation count $n_{\mathrm{conf}}$
\Ensure Imagination tree $\mathcal T$

\State Build a reference spine $\mathcal T$ from observations of $\tau$ at action-chunk boundaries
\State Cache the root world-model state; materialize and cache other spine states on first use
\State Mark the final spine node terminal and back up $y_\tau\in\{0,1\}$ along the spine
\State Set depth limit $D_\tau$ to the spine depth if $y_\tau=1$, and to $D_{\max}$ otherwise
\State Initialize $n_{\mathrm{sim}}\gets0$ and an empty queue $\mathcal Q_{\mathrm{mix}}$

\While{$n_{\mathrm{sim}}<B$ and $\mathcal T$ contains an expandable node}
    \State $v\gets\Call{SelectPUCTPW}{\mathcal T,c_{\mathrm{puct}},c_{\mathrm{pw}},\alpha}$
    \Comment{PUCT with progressive widening}
    \If{$v=\varnothing$}
        \State \textbf{break}
    \EndIf
    \State $n_{\mathrm{new}}\gets 2$ if $v$ has no imagined child; otherwise $n_{\mathrm{new}}\gets1$
    \For{$j=1,\ldots,n_{\mathrm{new}}$}
        \State $(\ell,\Delta n)\gets\Call{ExpandAndRollout}{v,D_\tau}$
        \State $n_{\mathrm{sim}}\gets n_{\mathrm{sim}}+\Delta n$
        \State Obtain $y_\ell\in\{0,1\}$ from the task-specific reward model
        \State Back up $y_\ell$ from leaf $\ell$ to the root:
        $N\gets N+1$, $W\gets W+y_\ell$, and $Q\gets W/N$
    \EndFor
    \State Recompute the relative bins of the imagined children of $v$
    \If{the children of $v$ occupy different bins}
        \State Push $v$ into $\mathcal Q_{\mathrm{mix}}$
    \EndIf

    \While{$\mathcal Q_{\mathrm{mix}}\neq\varnothing$ and $n_{\mathrm{sim}}<B$}
        \State Pop a parent $p$ from $\mathcal Q_{\mathrm{mix}}$
        \While{an expandable imagined child $u\in\mathcal C_{\mathrm{im}}(p)$ has
        $N(u)<n_{\mathrm{conf}}$ and $n_{\mathrm{sim}}<B$}
            \State Choose the least-visited such child
            $u\gets\arg\min_{u'}N(u')$
            \State Select an expandable descendant $v'$ within the subtree rooted at $u$
            \State $(\ell,\Delta n)\gets\Call{ExpandAndRollout}{v',D_\tau}$
            \State $n_{\mathrm{sim}}\gets n_{\mathrm{sim}}+\Delta n$
            \State Obtain $y_\ell$ from the reward model; back up $y_\ell$ and enqueue any new mixed group
        \EndWhile
    \EndWhile
\EndWhile
\State \Return $\mathcal T$
\end{algorithmic}
\end{algorithm}

Each call to \textsc{ExpandAndRollout} first restores $m_v$, samples an action
chunk from $\pi_\theta(\cdot\mid o_v,g,b=K)$ using a fresh noise seed, and
steps $\mathcal W_\phi$.  The resulting world-model state is cached at the new
child.  The highest-bin policy is then rolled out until success or $D_\tau$,
caching every intermediate node.  The routine returns the terminal leaf and
the number of simulated action chunks consumed.
In \textsc{SelectPUCTPW}, existing children use the PUCT score defined in the
main text with uniform sibling priors.  A node may contain at most
$\lceil c_{\mathrm{pw}}\max(1,N(v))^\alpha\rceil$ children.  Whenever widening
is permitted, a virtual child with zero empirical value and prior
$1/(|\mathcal C(v)|+1)$ competes with existing children; selecting it returns
$v$ for expansion.  Restoring $m_v$ before every expansion ensures that all
siblings share the same decision context without regenerating their common
trajectory prefix.  The budget $B$ counts simulated action chunks and is soft
because an initiated rollout and its atomic sibling pair are completed before
termination; truncating either would not yield comparable outcomes.  In our
configuration, the first imagined sibling
pair is always admitted, whereas later widening is restricted to mixed groups
and capped at a task-specific maximum width.

\begin{algorithm}[!t]
\caption{Extraction of tree-structured relative-value supervision}
\label{alg:rollback_labels}
\small
\begin{algorithmic}[1]
\Require Imagination tree $\mathcal T$; number of bins $K$
\Ensure Tree-derived training set $\mathcal D_{\mathrm{tree}}$
\State $\mathcal D_{\mathrm{tree}}\gets\varnothing$
\For{each parent $v$ with imagined children $\mathcal C_{\mathrm{im}}(v)$}
    \If{$|\mathcal C_{\mathrm{im}}(v)|<2$}
        \State \textbf{continue}
    \EndIf
    \For{each action-child pair $(\mathbf{a}_i,u_i)$ of $v$}
        \State $\hat{q}(v,\mathbf{a}_i)\gets
        |\mathcal L(u_i)|^{-1}\sum_{\ell\in\mathcal L(u_i)}y_\ell$
    \EndFor
    \State $q_{\max}(v)\gets\max_i\hat{q}(v,\mathbf{a}_i)$
    \If{$q_{\max}(v)=0$}
        \State \textbf{continue} \Comment{No successful continuation was observed}
    \EndIf
    \For{each $(\mathbf{a}_i,u_i)$}
        \State $\rho_i\gets\hat{q}(v,\mathbf{a}_i)/q_{\max}(v)$
        \State $b_i\gets\min\!\left(K,1+\lfloor K\rho_i\rfloor\right)$
    \EndFor
    \If{$|\{b_i\}_i|=1$}
        \State \textbf{continue} \Comment{Settled group; no relative preference signal}
    \EndIf
    \For{each $(\mathbf{a}_i,b_i)$}
        \State $\mathcal D_{\mathrm{tree}}\gets
        \mathcal D_{\mathrm{tree}}\cup\{(o_v,g,\mathbf{a}_i,b_i)\}$
    \EndFor
\EndFor
\State \Return $\mathcal D_{\mathrm{tree}}$
\end{algorithmic}
\end{algorithm}

\section{Experiment Details}
\label{app:details}

The main text describes the three tasks and the headline evaluation protocol.
Here, we provide the execution-level details needed for reproducibility.

\paragraph{Observation and Control Interface.}
All policies receive three RGB views (one head-mounted camera and two wrist
cameras) together with the robot's proprioceptive state.  Images from both
real execution and imagined rollouts are mapped to a common
$384\times512$ canvas before the policy-specific image transforms are applied.
The deployed action has 16 dimensions, comprising seven arm and one gripper
coordinate per side.  The low-level control loop runs at 30\,Hz.  For block
extraction and ball tossing, the policy predicts 50-step action chunks and is
executed with real-time chunking~\citep{rtc} using a delay of 6 steps and a replanning
stride of 32 steps.  The same observation processing, action interface, and
controller settings are used for all compared policies.

\paragraph{Trial Initialization and Termination.}
Before each trial, the robot is returned to a task-specific reset
configuration sampled from a pool constructed from the beginning of successful
demonstration segments.  This avoids evaluating every policy from a single
fixed arm pose while keeping the reset distribution identical across methods.
Object configurations are then randomized according to the task protocol.
Block-extraction episodes have a hard limit of 700 control steps
($23.3$\,s), whereas ball-tossing episodes are limited to 120 steps
($4.0$\,s).  An episode terminates early once the outcome is determined
or a safety stop is issued.  A tabletop-curling trial ends after
the human and robot have each completed two shots, and the winner is determined
from the final stone configuration.

\paragraph{Outcome Annotation.}
Real-robot trajectories are assigned binary success labels from their observed
physical outcomes.  Block extraction is successful only when the target block
is removed while the tower remains intact; a collapse or an unlabeled
termination is treated as failure.  Ball tossing is successful only when the
ball lands in the bucket, and all other terminations are failures.  Curling is
scored only after all four shots so that displacement by later shots is
accounted for.  The task-specific GE-Sim 2.0 world judge is used
to evaluate imagined world-model rollouts only; it does not determine the
reported real-world success rates.

\section{Hyperparameters}
\label{app:hparams}

Table~\ref{tab:app-hparams} lists the hyperparameters of \modelname{}; task-specific values are given as Block Extraction / Ball Tossing / Tabletop Curling, and all other values are shared across the three tasks.

\begin{table}[!htbp]
    \centering
    \caption{\textbf{Hyperparameters of \modelname{}.}}
    \label{tab:app-hparams}
    \small
    \setlength{\tabcolsep}{8pt}
    \begin{tabular}{llc}
        \toprule
        Component & Hyperparameter & Value \\
        \midrule
        \multirow{5}{*}{World model}
        & Initialization & GE-Sim 2.0 \\
        & Initial SFT (steps / GPUs / batch) & 10K / 32 / 128 \\
        & Initial DMD (steps / GPUs / batch) & 2,500 / 16 / 16 \\
        & Per adaptation & 400 LoRA SFT + 100 DMD steps \\
        & Frames per action chunk / frame rate & 25 / 16 fps \\
        \midrule
        \multirow{6}{*}{Policy}
        & Initialization & $\pi_{0.5}$ \\
        & Warm-up (steps / GPUs / batch) & 100K / 16 / 64 \\
        & Advantage bins $K$ & 5 \\
        & Condition dropout $p_{\mathrm{drop}}$ & 0.5 \\
        & Guidance strength $\beta$ & 1.5 \\
        & Action chunk / control frequency & 50 steps / 30 Hz \\
        \midrule
        \multirow{2}{*}{Reward model}
        & Initialization & GE-Sim 2.0 world judge \\
        & Success threshold & 0.5 \\
        \midrule
        \multirow{6}{*}{Search}
        & PUCT coefficient $c_{\mathrm{puct}}$ & 1.5 \\
        & Widening parameters $(c_{\mathrm{pw}}, \alpha)$ & (1.5, 0.5) / (4.0, 0.5) / (1.5, 0.5) \\
        & Actions per new branch & 2 \\
        & Confirmation count $n_{\mathrm{conf}}$ & 5 / 9 / 5 \\
        & Imagination budget $B$ (action chunks) & 256 \\
        & Maximum search depth $D_{\max}$ (action chunks) & 12 / 2 / 8 \\
        \midrule
        \multirow{2}{*}{System}
        & GPUs for search / policy training & 40 / 8 (H200) \\
        & Physical robots & 1 \\
        \bottomrule
    \end{tabular}
\end{table}

\section{Per-Position Results}
\label{app:per-position}

Tables~\ref{tab:app-block}--\ref{tab:app-curling} report success rates at each initial position.
Every policy is evaluated for 25 trials per position on Block Extraction and 15 trials per position on Ball Tossing and Tabletop Curling.
For Block Extraction, we additionally report the average number of blocks extracted in a row before failure; for Tabletop Curling, we report the average score obtained against the human opponent.

\begin{table}[!htbp]
    \centering
    \caption{\textbf{Per-position results on Block Extraction} (success rate, \%; 25 trials per position).}
    \label{tab:app-block}
    \small
    \setlength{\tabcolsep}{5pt}
    \begin{tabular*}{\linewidth}{@{\extracolsep{\fill}}lcccccccc}
        \toprule
        Method & P1 & P2 & P3 & P4 & P5 & P6 & Overall & Consec. Extr. \\
        \midrule
        $\pi_{0.5}$ + SFT & 64.00 & 80.00 & 44.00 & 40.00 & 44.00 & 20.00 & 48.67 & 1.60 \\
        $\pi_{0.5}$ + RECAP & 48.00 & 84.00 & 40.00 & 48.00 & 72.00 & 44.00 & 56.00 & 1.78 \\
        Frozen World Model & 72.00 & 76.00 & 48.00 & 68.00 & 60.00 & 64.00 & 64.67 & 1.90 \\
        Direct Reward Supervision & 76.00 & 64.00 & 48.00 & 64.00 & 64.00 & 40.00 & 59.33 & 1.73 \\
        $\pi_{0.5}$ + RISE w/ GE-Sim2 & 60.00 & 80.00 & 32.00 & 68.00 & 52.00 & 60.00 & 58.67 & 2.00 \\
        \midrule
        \textbf{\modelname{} (Ours)} & 72.00 & 56.00 & 56.00 & 64.00 & 84.00 & 84.00 & \textbf{69.33} & \textbf{2.60} \\
        \bottomrule
    \end{tabular*}
\end{table}

\begin{table}[!htbp]
    \centering
    \caption{\textbf{Per-position results on Ball Tossing} (success rate, \%; 15 trials per position).}
    \label{tab:app-throw}
    \small
    \setlength{\tabcolsep}{6pt}
    \begin{tabular*}{\linewidth}{@{\extracolsep{\fill}}lccccccc}
        \toprule
        Method & P1 & P2 & P3 & P4 & P5 & P6 & Overall \\
        \midrule
        $\pi_{0.5}$ + SFT & 13.33 & 60.00 & 0.00 & 0.00 & 46.67 & 20.00 & 23.33 \\
        $\pi_{0.5}$ + RECAP & 0.00 & 60.00 & 20.00 & 13.33 & 46.67 & 13.33 & 25.56 \\
        Frozen World Model & 0.00 & 46.67 & 26.67 & 13.33 & 53.33 & 20.00 & 26.67 \\
        $\pi_{0.5}$ + RISE w/ GE-Sim2 & 13.33 & 73.33 & 13.33 & 13.33 & 20.00 & 40.00 & 28.89 \\
        \midrule
        \textbf{\modelname{} (Ours)} & 26.67 & 60.00 & 33.33 & 46.67 & 53.33 & 20.00 & \textbf{40.00} \\
        \bottomrule
    \end{tabular*}
\end{table}

\begin{table}[!htbp]
    \centering
    \caption{\textbf{Per-position results on Tabletop Curling} (success rate, \%; 15 trials per position).}
    \label{tab:app-curling}
    \small
    \setlength{\tabcolsep}{3pt}
    \begin{tabular*}{\linewidth}{@{\extracolsep{\fill}}lccccccccccc}
        \toprule
        Method & P1 & P2 & P3 & P4 & P5 & P6 & P7 & P8 & P9 & Overall & Score \\
        \midrule
        $\pi_{0.5}$ + SFT & 60.00 & 86.67 & 33.33 & 26.67 & 60.00 & 0.00 & 40.00 & 80.00 & 53.33 & 48.89 & 0.35 \\
        $\pi_{0.5}$ + RECAP & 60.00 & 66.67 & 13.33 & 26.67 & 60.00 & 0.00 & 60.00 & 73.33 & 100.00 & 51.11 & 0.44 \\
        Frozen World Model & 53.33 & 53.33 & 13.33 & 6.67 & 33.33 & 46.67 & 80.00 & 100.00 & 86.67 & 52.59 & 0.51 \\
        $\pi_{0.5}$ + RISE w/ GE-Sim2 & 20.00 & 86.67 & 20.00 & 26.67 & 40.00 & 80.00 & 20.00 & 80.00 & 100.00 & 52.59 & 0.43 \\
        \midrule
        \textbf{\modelname{} (Ours)} & 86.67 & 80.00 & 60.00 & 20.00 & 20.00 & 80.00 & 80.00 & 80.00 & 73.33 & \textbf{64.44} & \textbf{0.79} \\
        \bottomrule
    \end{tabular*}
\end{table}

\section{Additional Qualitative Results}
\label{app:qualitative}

\begin{figure}[!htbp]
  \centering
  \includegraphics[width=\linewidth]{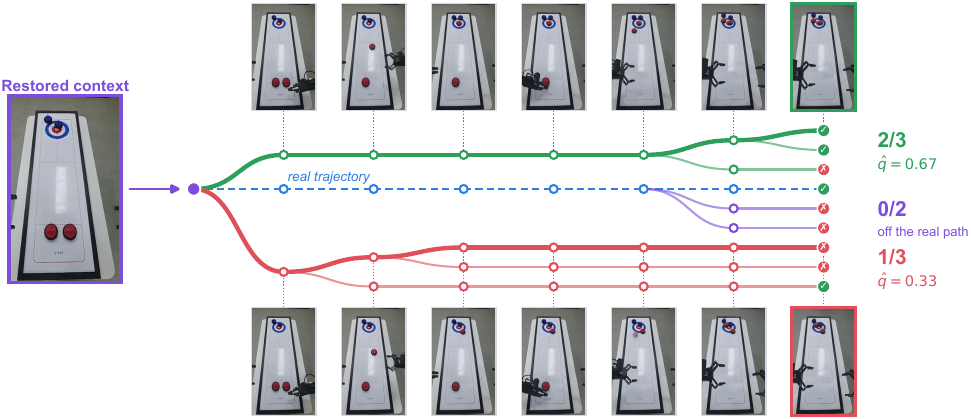}
\caption{
\textbf{Search tree on Tabletop Curling with recorded leaf outcomes.} Besides branching from the restored context, \modelname rolls back to an intermediate decision point on the real trajectory and explores alternative continuations from there (purple; here color marks where a branch starts, not its outcome).
}
  \label{fig:app-tree-curling}
\end{figure}

\begin{figure}[!htbp]
  \centering
  \includegraphics[width=\linewidth]{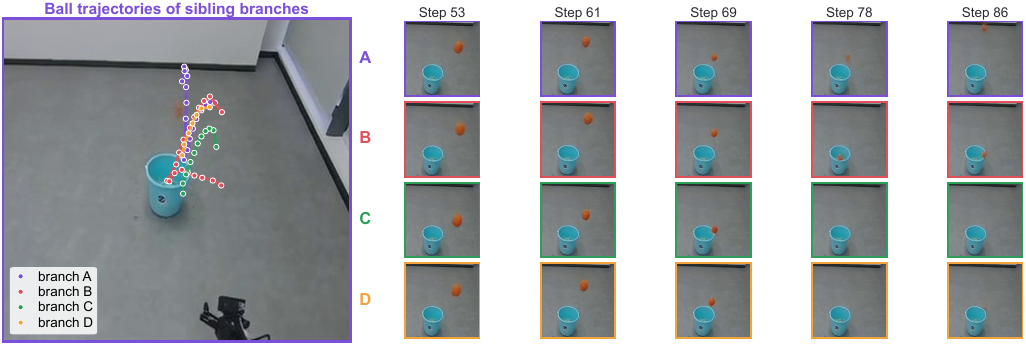}
\caption{
\textbf{Sibling continuations on Ball Tossing.} Left: ball trajectories of four sibling branches expanded from the same context. Right: close-ups around the bucket for each branch.
}
  \label{fig:app-throw}
\end{figure}

\paragraph{Search Tree on Tabletop Curling.}
Figure~\ref{fig:app-tree-curling} shows another search tree with its recorded leaf outcomes. The two root subtrees attain 2/3 and 1/3 successful continuations, yielding subtree values of 0.67 and 0.33, while branches rolled back from the real trajectory probe a late decision point.

\paragraph{Sibling Continuations on Ball Tossing.}
Figure~\ref{fig:app-throw} compares four sibling branches expanded from the same context. Although all branches share the same history, the sampled action chunks produce distinct ball trajectories, illustrating the diversity of explored counterfactual continuations.

\begin{figure}[!htbp]
  \centering
  \includegraphics[width=\linewidth]{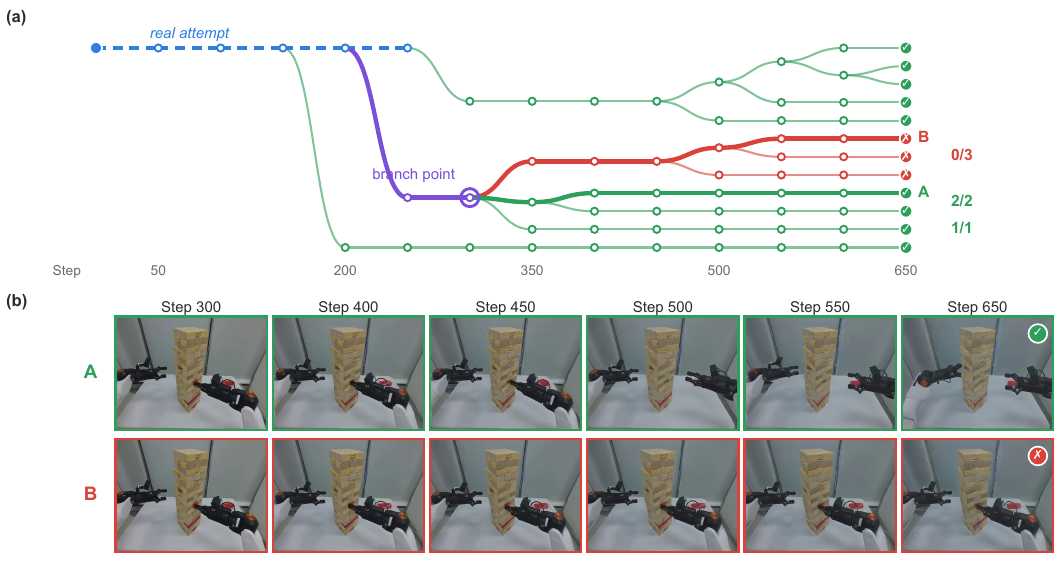}
\caption{
\textbf{Search tree on Block Extraction.} (a) Search tree with leaf outcomes, read from the imagined head-camera videos. The blue dashed path is a real attempt and the other paths are imagined continuations; edges are green, red, or purple when the continuations below all succeed, all fail, or are mixed. The ring marks a branch point inside imagination at step 300, whose three children reach 0/3, 2/2, and 1/1 successful continuations. (b) Continuations A and B from this branch point at the same control steps (30\,Hz): in A the gripper pulls the red block out of the tower and withdraws with it, while in B it is still in the tower at the end of the horizon.
}
  \label{fig:app-jenga-tree}
\end{figure}

\paragraph{Search Tree on Block Extraction.}
Figure~\ref{fig:app-jenga-tree} shows a search tree on Block Extraction with the outcome of each imagined continuation. \modelname rolls back to several points of a real attempt and expands 12 imagined continuations, 9 of which extract the red block with the tower intact. Branching also happens inside imagination: from the same imagined state at step 300, three sampled action chunks lead to subtrees with 0/3, 2/2, and 1/1 successful continuations. Panel (b) compares one continuation from each of the first two subtrees. Both start from the identical state, but in A the gripper pulls the red block out by step 500, whereas in B it stays in the tower until step 650. The resulting subtree values (0 vs.\ 1) single out the chunk that leads to extraction, which is the relative signal on which the policy is trained.

\begin{figure}[!htbp]
  \centering
  \includegraphics[width=\linewidth]{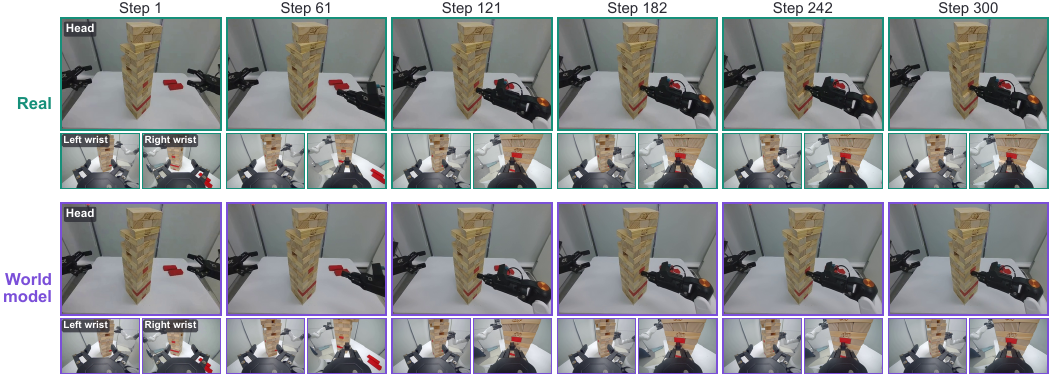}
\caption{
\textbf{Open-loop world-model predictions on Block Extraction.} Real recording (top) and world-model prediction under the same recorded actions (bottom) for one episode, from the head and both wrist cameras. Steps are control steps at 30\,Hz.
}
  \label{fig:app-jenga-fidelity}
\end{figure}

\paragraph{World-Model Fidelity on Block Extraction.}
Figure~\ref{fig:app-jenga-fidelity} visualizes one episode of the open-loop evaluation in Section~\ref{sec:fidelity}. Under the recorded actions, the world model reproduces the arm approaching the tower and the gripper closing on the red block over 300 control steps, consistently across the head and both wrist cameras, and keeps the tower intact as in the real recording.

\begin{figure}[!htbp]
  \centering
  \includegraphics[width=\linewidth]{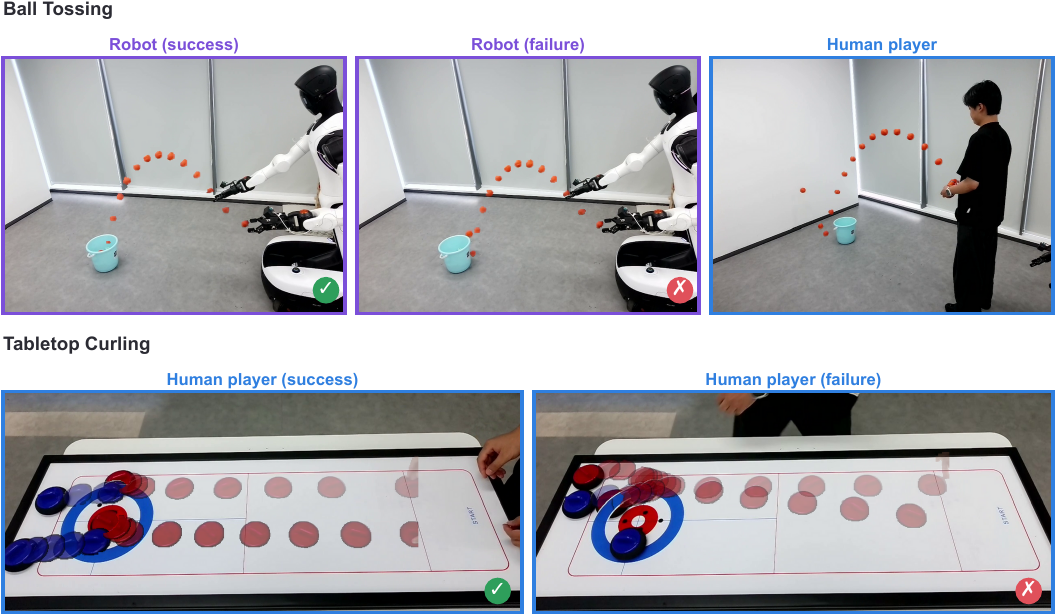}
  \captionsetup{format=loose}
\caption{
\textbf{Real-world executions.} Stroboscopic composites overlay the ball (top) and sliding stones (bottom) over time onto a single frame. Top: ball tossing by the robot and by a human player. Bottom: tabletop curling by human players, with successful and failed deliveries.
}
  \label{fig:app-realworld}
\end{figure}

\paragraph{Real-World Executions.}
Figure~\ref{fig:app-realworld} visualizes real-world executions with stroboscopic composites. In ball tossing, successful and failed robot throws differ mainly in the release that determines the landing point, with failures bouncing off the rim. In tabletop curling, successful human deliveries bring the red stones to rest in the house while displacing the opponent's stones, whereas failed deliveries leave them outside the scoring area.

\begin{figure}[!htbp]
  \centering
  \includegraphics[width=\linewidth]{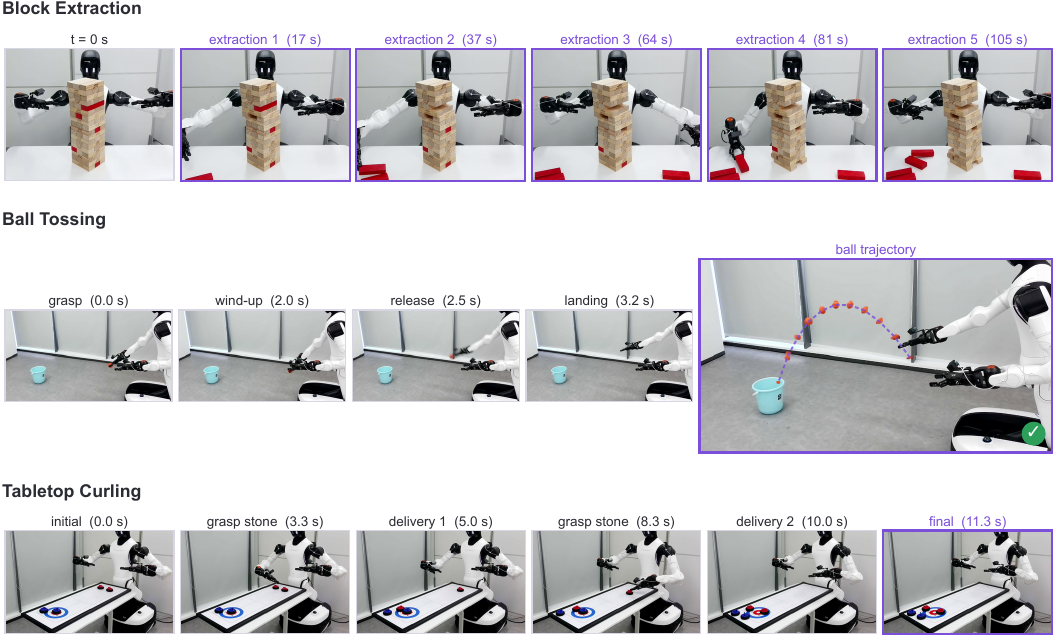}
\caption{
\textbf{Third-person views of \modelname{} on the real robot.} Top: five consecutive extractions without collapsing the tower. Middle: a successful toss, with the ball trajectory overlaid on the final frame. Bottom: two stone deliveries that leave the red stones in the house.
}
  \label{fig:app-thirdperson}
\end{figure}

\paragraph{Third-Person Executions.}
Figure~\ref{fig:app-thirdperson} shows real-robot executions of the final \modelname{} policy: five consecutive block extractions, a successful toss, and two curling deliveries.

\section{Failure Modes}
\label{app:failures}

\begin{figure}[!htbp]
  \centering
  \includegraphics[width=\linewidth]{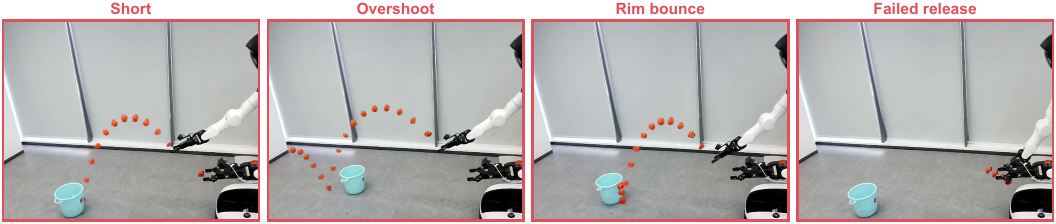}
\caption{
\textbf{Typical failures on Ball Tossing.} Stroboscopic composites of unsuccessful throws sampled from real-robot data collected during training.
}
  \label{fig:app-failures}
\end{figure}

Figure~\ref{fig:app-failures} shows the four failure types we observe on Ball Tossing.
Short and overshooting throws come from errors in release velocity, and rim bounces from small errors in release direction; these three types account for most failures.
Failed releases, where the gripper opens too late and the ball drops near the robot, are less frequent.
All four types are decided within a short window around release, so a single action chunk largely determines the outcome of the whole episode.

\section{Limitations and Future Work}
\label{app:limitations}
\modelname relies on a task-specific reward model and on world-model fidelity over long contact-rich horizons.
The fidelity analysis in Section~\ref{sec:fidelity} shows that prediction error grows along a rollout, and each policy in this work is trained for a single task with one physical robot.
Future work includes learning outcome detectors jointly with the world model and extending \modelname{} to multi-task policies.
Another direction is to scale the system to multiple robots that share one search cluster.